%% file: icml.tex
\documentclass{article}

\usepackage{microtype}
\usepackage{graphicx}
\usepackage{subcaption}
\usepackage{booktabs} % for professional tables
\usepackage{multirow}   % 支持多行合并的单元格
\usepackage{makecell} 
\usepackage{adjustbox}
\usepackage{pifont}
\usepackage{bbm}
\usepackage{enumitem}
\usepackage[most]{tcolorbox}

\usepackage{hyperref}

\usepackage[preprint]{icml2026}

\usepackage{amsmath}
\usepackage{amssymb}
\usepackage{mathtools}
\usepackage{amsthm}

\usepackage[capitalize,noabbrev]{cleveref}

\theoremstyle{plain}

\theoremstyle{definition}

\theoremstyle{remark}

\usepackage[textsize=tiny]{todonotes}

\icmltitlerunning{Chatting with Images for Introspective Visual Thinking}

\begin{document}

\twocolumn[
  \icmltitle{Chatting with Images for Introspective Visual Thinking}



  \icmlsetsymbol{equal}{*}
  \icmlsetsymbol{cor}{$\dagger$}

  \begin{icmlauthorlist}
    \icmlauthor{Junfei Wu}{equal,nlpr,ucas}
    \icmlauthor{Jian Guan}{equal,ant}
    \icmlauthor{Qiang Liu}{nlpr,ucas}
    \icmlauthor{Shu Wu}{cor,nlpr,ucas}
    \icmlauthor{Liang Wang}{nlpr,ucas}
    \icmlauthor{Wei Wu}{cor,ant}
    \icmlauthor{Tieniu Tan}{nlpr,ucas,nju}
  \end{icmlauthorlist}
  
  \icmlaffiliation{nlpr}{NLPR, MAIS, Institute of Automation, Chinese Academy of Sciences}
  \icmlaffiliation{ucas}{University of Chinese Academy of Sciences}
  \icmlaffiliation{ant}{Ant Group}
  \icmlaffiliation{nju}{Nanjing University}

  \icmlcorrespondingauthor{Shu Wu}{shu.wu@nlpr.ia.ac.cn}
  \icmlcorrespondingauthor{Wei Wu}{wuwei19850318@gmail.com}

  \icmlkeywords{Machine Learning, ICML}
    
  \vskip 0.3in
]



\printAffiliationsAndNotice{\icmlEqualContribution}

\input{sec/0_abstract}

\input{sec/1_intro}
\input{sec/2_related_works}
\input{sec/3_method}
\input{sec/4_experiment}

\input{sec/5_conclusion}

\section*{Impact Statement}

This work aims to improve the reliability and reasoning capability of large vision--language models by enabling language-guided, multi-turn interaction with visual evidence. If deployed responsibly, such models may benefit applications that require careful visual grounding, such as grounded question answering, education, and scientific or engineering assistance.

% Potential negative impacts mirror those of general-purpose vision--language systems. More capable visual reasoning could be misused for generating misleading content or for automating harmful decision-making in high-stakes settings. In addition, the proposed iterative region selection and re-encoding mechanism may amplify existing dataset biases by focusing attention on spurious cues, and its increased inference-time computation could raise energy use at scale.

% We do not introduce new data sources containing personal information, and our approach is intended as a methodological contribution rather than a deployment-ready system. We encourage future work to evaluate fairness and robustness across demographic and cultural contexts, to add safeguards for high-stakes use, and to further quantify and reduce computational overhead.


\bibliography{icml}
\bibliographystyle{icml2026}

%%%%%%%%%%%%%%%%%%%%%%%%%%%%%%%%%%%%%%%%%%%%%%%%%%%%%%%%%%%%%%%%%%%%%%%%%%%%%%%
%%%%%%%%%%%%%%%%%%%%%%%%%%%%%%%%%%%%%%%%%%%%%%%%%%%%%%%%%%%%%%%%%%%%%%%%%%%%%%%
% APPENDIX
%%%%%%%%%%%%%%%%%%%%%%%%%%%%%%%%%%%%%%%%%%%%%%%%%%%%%%%%%%%%%%%%%%%%%%%%%%%%%%%
%%%%%%%%%%%%%%%%%%%%%%%%%%%%%%%%%%%%%%%%%%%%%%%%%%%%%%%%%%%%%%%%%%%%%%%%%%%%%%%
\newpage
\appendix
% \onecolumn

\input{sec/6_suppl}


% You can have as much text here as you want. The main body must be at most $8$
% pages long. For the final version, one more page can be added. If you want, you
% can use an appendix like this one.

% The $\mathtt{\backslash onecolumn}$ command above can be kept in place if you
% prefer a one-column appendix, or can be removed if you prefer a two-column
% appendix.  Apart from this possible change, the style (font size, spacing,
% margins, page numbering, etc.) should be kept the same as the main body.
%%%%%%%%%%%%%%%%%%%%%%%%%%%%%%%%%%%%%%%%%%%%%%%%%%%%%%%%%%%%%%%%%%%%%%%%%%%%%%%
%%%%%%%%%%%%%%%%%%%%%%%%%%%%%%%%%%%%%%%%%%%%%%%%%%%%%%%%%%%%%%%%%%%%%%%%%%%%%%%

\end{document}

%% file: sec/0_abstract.tex
\begin{abstract}

Current large vision-language models (LVLMs) typically rely on text-only reasoning based on a single-pass visual encoding, which often leads to loss of fine-grained visual information. Recently the proposal of ``thinking with images'' attempts to alleviate this limitation by manipulating images via external tools or code; however, the resulting visual states are often insufficiently grounded in linguistic semantics, impairing effective cross-modal alignment - particularly when visual semantics or geometric relationships must be reasoned over across distant regions or multiple images. To address these challenges, we propose {``chatting with images''}, a new framework that reframes visual manipulation as language-guided feature modulation. Under the guidance of expressive language prompts, the model dynamically performs joint re-encoding over multiple image regions, enabling tighter coupling between linguistic reasoning and visual state updates. We instantiate this paradigm in \textbf{\textsc{ViLaVT}}, a novel LVLM equipped with a dynamic vision encoder explicitly designed for such interactive visual reasoning, and trained it with a two-stage curriculum combining supervised fine-tuning and  reinforcement learning to promote effective reasoning behaviors. Extensive experiments across eight benchmarks demonstrate that \textbf{\textsc{ViLaVT}} achieves strong and consistent improvements, with particularly pronounced gains on complex multi-image and video-based spatial reasoning tasks. Code and model are
available\footnote{https://github.com/AntResearchNLP/ViLaVT}.
\end{abstract}

%% file: sec/1_intro.tex
\section{Introduction}
\label{sec:intro}

Large Language Models (LLMs) have established new frontiers in complex problem-solving with long-horizon cognitive deliberation~\citep{openai2024o1preview,guo2025deepseek}, demonstrating impressive abilities in domains like mathematics \citep{wei2022chain,lightman2023let} and software engineering \citep{chen2021evaluatinglargelanguagemodels,hou2024large}.
%A key driver of this success has been the adoption of long-horizon reasoning, which allows for extended, step-by-step cognitive deliberation~\citep{openai2024o1preview,guo2025deepseek}. 
% These advances have naturally spurred efforts to imbue 
This has motivated to imbue Large Vision-Language Models (LVLMs) with similar cognitive capabilities, aiming to unlock sophisticated multimodal reasoning \citep{yang2025r1,openai2025o3o4,wu2025reinforcing}. 
Despite this progress, the dominant multimodal reasoning paradigm largely adheres to a “thinking about images” workflow~\citep{su2025thinking} and remains bottlenecked by single-pass visual encoding: models first produce a fixed set of visual tokens and then conduct subsequent reasoning primarily in the language space~\citep{yang2025r1,hong2025glm}. This design implicitly assumes that such static visual tokens can preserve and be fully exploited for rich visual semantics~\citep{huh2024position}. In practice, however, 
% task-relevant details can be lost during compression into static visual embeddings and further diluted as reasoning proceeds purely in the textual domain
task-relevant details can be lost during this compression and further attenuated as reasoning proceeds purely in the textual domain~\citep{huh2024position,li2024multimodal,wu2025reinforcing}. Moreover, this limitation is more pronounced in multi-image spatial reasoning: the common practice of encoding each image independently and aggregating them only in the textual domain obscures geometric relationships across views~\citep{wang2025vggt}.

% Despite this progress, the dominant multimodal reasoning paradigm remains fundamentally limited. Specifically, current LVLMs largely adhere to a ``thinking about images'' paradigm~\citep{su2025thinking}: they perform a one-time encoding of the visual input to generate static embeddings, upon which all subsequent reasoning unfolds exclusively in the textual domain~\citep{yang2025r1,hong2025glm}. This approach relies on a flawed premise that rich visual semantics can be losslessly compressed into textual representations~\citep{huh2024position}, a process that in practice inevitably discards nuanced details~\citep{li2024multimodal,wu2025reinforcing}. For example, the failure is especially pronounced in multi-image contexts for spatial relationships, as the common practice of encoding multiple images in isolation inherently severs the geometric relationships between views~\citep{wang2025vggt}.

\begin{figure*}
    \centering
    \includegraphics[width=0.9\textwidth]{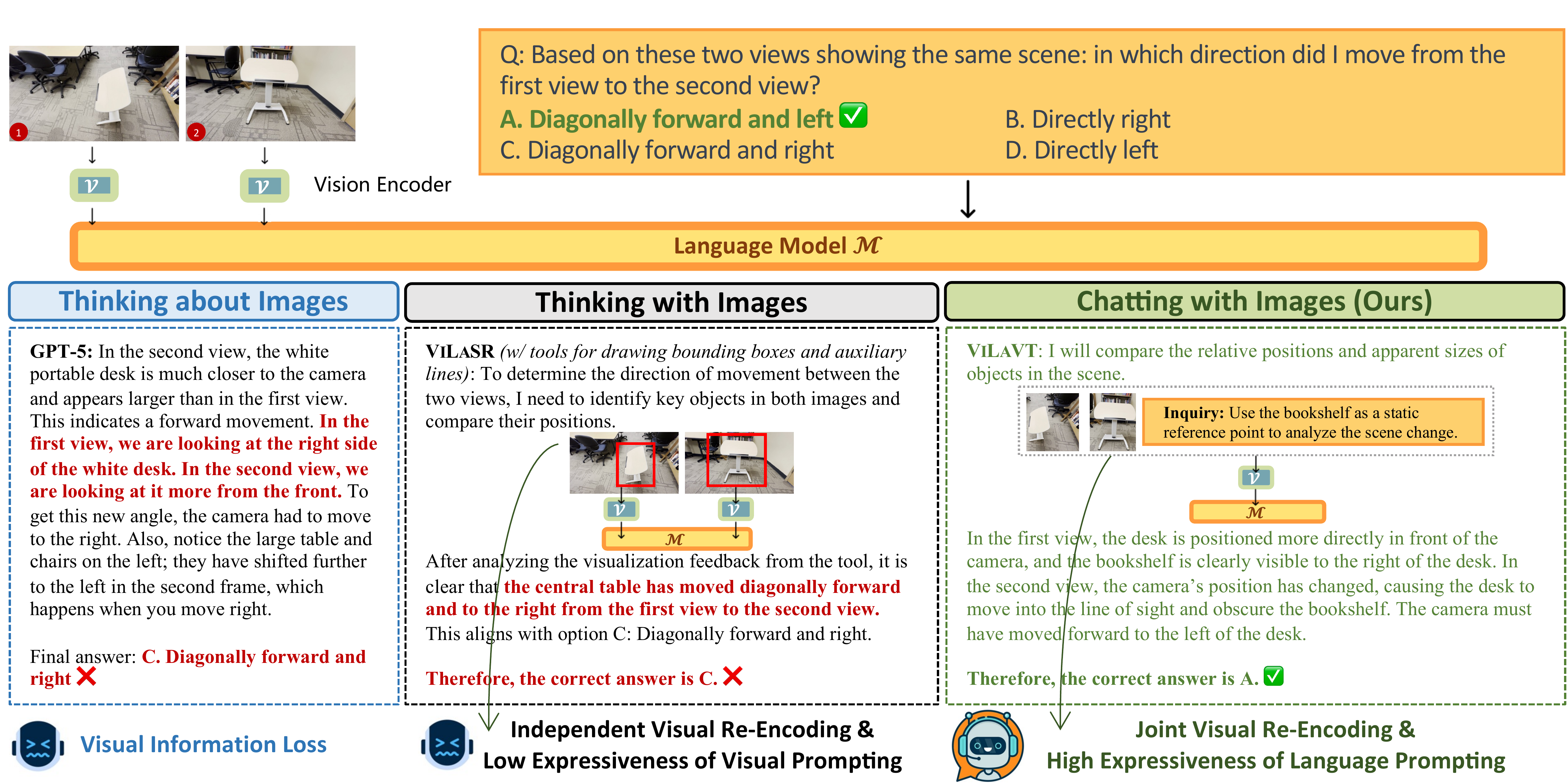}
    \caption{A qualitative comparison of three reasoning paradigms on a multi-view spatial reasoning task.
    \textbf{``Thinking about Images'' (Left):} A static LVLM relies on a one-time visual encoding, resulting in information loss. This leads to a flawed understanding of the objects' spatial relationships and an incorrect answer.
    \textbf{``Thinking with Images'' (Center):} This method invokes an external tool to highlight salient information while re-encoding each image independently. However, this visual prompting technique lacks expressiveness and fails to convey the necessary cognitive intent, resulting in a flawed comparison.
    \textbf{``Chatting with Images'' (Right):} Our model, in contrast, leverages language prompting. The generated inquiry expresses a high-level cognitive intent. This highly expressive, declarative prompt guides a joint visual re-encoding of both images, enabling a relational comparison at the feature level and leading to the correct inference. %, demonstrating the power and precision of our versatile reasoning paradigm.
}

    \label{fig:example}
\end{figure*}

To transcend this limitation, recent work has advocated for a paradigm shift towards ``thinking with images'', where models interleave textual reasoning with targeted visual manipulations and iteratively re-encode visual inputs to enrich the context with previously lost details~\citep{openai2025o3o4}. The attempts in this direction primarily focus on two categories of manipulation: (1) Tool-based manipulation, where the model invokes predefined tools (e.g., image rotation)~\citep{cheng2024least,qi2025cogcom,wu2025reinforcing}; and (2) Programmatic manipulation, where the model generates executable image processing code~\citep{openai2025o3o4,zhang2025thyme}. 
Despite this conceptual advance, most existing methods still primarily target single-image manipulation. While recent approaches (e.g., \textsc{ViLaSR}~\citep{wu2025reinforcing}) can manipulate multiple images within a single reasoning episode, they often encode each view independently and fuse views only the textual domain, making cross-view geometric alignment difficult. Moreover, the manipulation interface is typically limited to low-level pixel/geometry operations~\citep{wu2024visualpromptingmultimodallarge}, which makes it nontrivial to express high-level relational intent. 
Considering a task like “Find the car that is parked incorrectly”, expressing such high-level relational intent via low-level operations is often cumbersome, requiring a brittle and inefficient chain of operations for detecting cars, lines, and comparing relations. 
% Therefore, a core challenge persists: 
The key challenge is thus to devise a visual manipulation mechanism that is both scalable and expressive, seamlessly bridging high-level intent with low-level feature modulation.

To address these challenges, we propose {``chatting with images,''} a versatile reasoning paradigm that reframes visual manipulation as language-guided feature modulation (Figure~\ref{fig:example}). 
At each reasoning step, the model predicts a set of relevant regions of interest and formulates a natural language inquiry, which serves as a highly expressive {language prompt}. This declarative prompt leverages the compositional power of language to express complex cognitive intent (e.g., ``Use the bookshelf as a static reference'').
Instead of triggering external tools, this inquiry directly conditions {dynamic, joint encoding} over these regions. This process facilitates the model to retrieve fine-grained cues and perform cross-view interactions directly in feature space, thereby alleviating the limitations of independently encoding each view.
By leveraging language as a unified control interface, our paradigm reduces reliance on handcrafting a vast combinatorial of specialized tools or program chains.

% ========== original ==========
% To address these challenges, we propose {``chatting with images,''} a versatile reasoning paradigm that reframes visual manipulation as language-guided feature modulation (Figure~\ref{fig:example}). At each reasoning step, the model formulates a natural language inquiry, which serves as a highly expressive {language prompt}. This declarative prompt leverages the compositional power of language to express complex cognitive intent (e.g., ``Use the bookshelf as a static reference''), overcoming the rigidity and low expressiveness of procedural visual prompts. Crucially, instead of triggering external tools, this inquiry directly guides a {dynamic, joint re-encoding} of features from multiple image regions. This process overcomes the information loss from independent encoding by enabling deep, relational comparisons at the feature level. By leveraging the model's intrinsic language capabilities, this paradigm provides a unified and scalable solution, obviating the need to handcraft a vast combinatorial of specialized tools or code libraries.

 %This elegant architecture fosters a truly generalizable and powerful multimodal reasoning framework.

% To realize this paradigm, we develop \textbf{\textsc{ViLaVT}},a novel \underline{\textbf{\textsc{V}}}ision-\underline{\textbf{\textsc{L}}}anguage model designed to enable a new form of \underline{\textbf{\textsc{V}}}isual \underline{\textbf{\textsc{T}}}hinking through two core innovations. 

To realize this paradigm, we develop \textbf{\textsc{ViLaVT}}, a novel \underline{\textbf{\textsc{V}}}ision-\underline{\textbf{\textsc{L}}}anguage model that redefines human-like \underline{\textbf{\textsc{V}}}isual \underline{\textbf{\textsc{T}}}hinking. 
Instead of generating external commands for discrete image operations, \textsc{ViLaVT} learns to introspectively modulate its own visual perception. To this end, we introduce a dynamic vision encoder designed to support this introspective capability. We re-architect the vanilla vision encoder to jointly process multiple, non-contiguous images conditioned on a textual inquiry, enabling cross-view interactions within the encoder instead of late fusion in the textual domain.
% transforming manipulation from an external, non-differentiable step into a continuous, internal feature modulation. 
For training, we first apply supervised fine-tuning (SFT) on a hybrid corpus combining repurposed trajectories from prior tool-/code-based methods with newly synthesized reasoning traces.
% For training, we first employ Supervised Fine-Tuning (SFT), bootstrapping learning on a hybrid dataset comprising both programmatically repurposed data from prior tool- or code-based methods and newly synthesized reasoning trajectories.
Following SFT, Reinforcement Learning (RL) elevates the model from an imitator to a strategic problem-solver, enabling it to explore the reasoning space and discover more effective solution pathways.

% ========== original ==========
% To realize this paradigm, we develop \textbf{\textsc{ViLaVT}}, a novel \underline{\textbf{\textsc{V}}}ision-\underline{\textbf{\textsc{L}}}anguage model that redefines human-like \underline{\textbf{\textsc{V}}}isual \underline{\textbf{\textsc{T}}}hinking. 
% Instead of generating external commands to manipulate images at the pixel level, \textsc{ViLaVT} learns to introspectively modulate its own visual perception. To this end, we introduce a dynamic vision encoder designed to support this introspective capability. We re-architect the vanilla vision encoder to jointly process multiple, non-contiguous images conditioned on a textual inquiry, transforming manipulation from an external, non-differentiable step into a continuous, internal feature modulation. For training, we first employ Supervised Fine-Tuning (SFT), bootstrapping learning on a hybrid dataset comprising both programmatically repurposed data from prior tool- or code-based methods and newly synthesized reasoning trajectories.
% Following SFT, Reinforcement Learning (RL) elevates the model from an imitator to a strategic problem-solver, enabling it to explore the reasoning space and discover more effective solution pathways.

Extensive experiments show that \textsc{ViLaVT} improves performance over prior paradigms, achieving state-of-the-art results on \textbf{5 out of 8} benchmarks with especially strong gains on multi-image and video-based spatial reasoning. Our ablations disentangle the contributions of the reasoning paradigm, vision encoder, and staged training, revealing a monotonic performance lift as these components are progressively integrated.
Furthermore, controlled analyses under information-scarce settings show improved detail preservation and stronger cross-view integration, resulting in a \textbf{24.8\%} performance gain over a vanilla encoder.
% Extensive experiments show that \textsc{ViLaVT} substantially improves performance over prior paradigms. It achieves state-of-the-art results on \textbf{5 out of 8} benchmarks, with particularly strong gains on complex multi-image and video-based spatial reasoning tasks. Moreover, comparisons against an ablated \textsc{ViLaVT} with a vanilla vision encoder highlight the benefit of our design for inquiry-conditioned joint encoding. Controlled analyses further indicate that the dynamic vision encoder remains effective in high-resolution and cross-view scenes, especially under information-scarce settings. Visualizations of attention patterns suggest that our approach helps the vision encoder focus on task-relevant fine-grained details. These findings help explain \textsc{ViLaVT}'s strong performance across diverse evaluations. 
In summary, our main contributions are:
\begin{itemize}[leftmargin=12pt, topsep=2pt, itemsep=0pt]
    \item We introduce {``chatting with images,''} a new reasoning paradigm that reframes visual manipulation as language-guided feature modulation, which mitigates the information loss of static models and the low expressiveness of tool/code-based visual prompting techniques.
    \item We develop {\textsc{ViLaVT}}, a novel LVLM featuring a dynamic vision encoder designed to support this paradigm. The encoder is architected to jointly process multiple images conditioned on a language inquiry. 
    \item We demonstrate strong empirical results on 8 benchmarks, achieving state-of-the-art performance on \textbf{5/8} of them and consistent improvements across the rest, especially on multi-image and video-based spatial reasoning.
\end{itemize} 

%% file: sec/2_related_works.tex
\begin{figure*}
    \centering
    \includegraphics[width=0.95\textwidth]{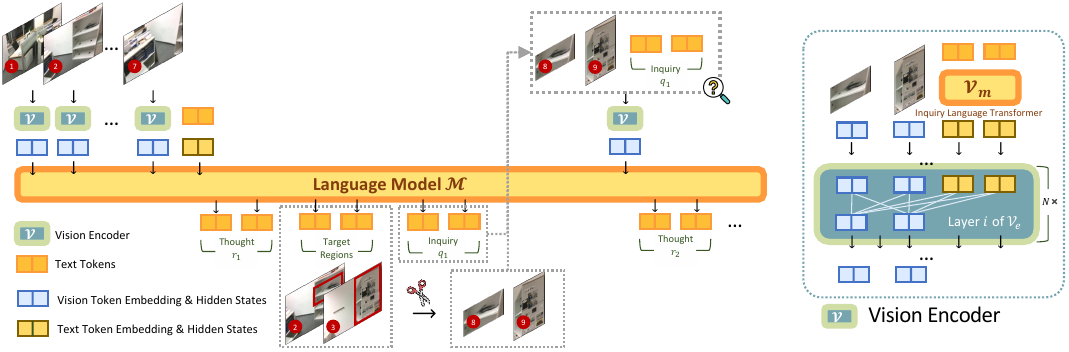}
    \caption{\textbf{Left}: The iterative reasoning process of \textsc{ViLaVT}; \textbf{Right:} the architecture of the dynamic vision encoder. The ``chatting with image'' reasoning paradigm unfolds as:
    % Our model unifies visual reasoning into an end-to-end, auto-regressive loop.
\textbf{(1) Initial Encoding:} All input images/frames are initially encoded independently into vision token embeddings.
\textbf{(2) Stepwise Reasoning:} The language model generates a triplet $s_t=(r_t, q_t, z_t)$, i.e., an internal thought, a natural language inquiry for visual re-encoding, and a set of target regions.
\textbf{(3) Targeted Re-encoding:} Our dynamic vision encoder \textbf{(Right)} takes the textual inquiry $q_t$ and the specified visual regions (cropped and upscaled from source images/frames) as input, which employs a hybrid attention strategy to jointly process vision and text tokens, producing re-encoded vision token embeddings.
\textbf{(4) Iteration:} These newly generated vision token embeddings are then passed back to the language model, enriching its context and enabling it to generate the next, more informed reasoning ($s_{t+1}$). This iteration continues until a final answer is reached.}
    \label{fig:hyperPara}
\end{figure*}
\section{Related Works}
\label{sec:related_works}

%Recent advancements in LVLMs have significantly advanced the frontier of multimodal reasoning. 
The evolution of visual reasoning models can be broadly categorized into two dominant paradigms: ``thinking about images,'' which treats reasoning as a post-perception textual process, and the emerging ``thinking with images,'' which reconceptualizes vision as an interactive component of the reasoning loop itself. In this section, we review key works within both paradigms.

\subsection{Thinking {about} Images}

Most state-of-the-art LVLMs~\citep{liu2024llavanext,li2024llavaonevisioneasyvisualtask,bai2025qwen2} operate under the ``thinking {about} images'' paradigm. Architecturally, these models process an image once with a powerful visual encoder~\citep{dosovitskiy2021an,tschannen2025siglip,fini2025multimodal} to generate a set of visual embeddings. To bridge the modality gap, these visual embeddings are projected into the language embedding space, typically through a bottleneck mechanism like a small number of learnable query tokens~\citep{li2023blip} or a simple linear projection layer~\citep{liu2024visual}. The resulting features are then treated as a static ``visual prefix'' for an LLM, which performs all subsequent reasoning. The LVLMs' capabilities can be improved through extensive training phases, including supervised fine-tuning (SFT) on instruction-following datasets~\citep{liu2024visual}, reinforcement learning from human feedback~(RLHF) to align with human preferences~\citep{sun2023aligning,yu2024rlhf,zhang2025mmrlhf}, and reinforcement learning with verifiable reward (RLVR) to enhance the reliability of long-horizon reasoning~\citep{yang2025r1,hong2025glm, feng2025video, feng2025onethinker}. Despite their impressive performance, the core bottleneck of this paradigm is the irreversible loss of information. This is rooted in the aggressive compression required to map high-dimensional, noisy visual data into a manageable sequence of tokens for the LLM, inevitably leading to the loss of fine-grained details and spatial nuances before any reasoning occurs.
\subsection{Thinking {with} Images}
More recent work has begun to explore the ``thinking {with} images'' paradigm~\citep{openai2025o3o4,su2025thinking}, aiming to create an interactive reasoning process. These approaches primarily manifest in three categories: \textbf{(1) Tool-based manipulation,} where the model dynamically invokes external tools to manipulate the visual input during reasoning~\citep{cheng2024least,wu2025reinforcing,zheng2025deepeyes,su2025openthinkimg}. However, the black-box nature of the tools prevents end-to-end optimization with the model, and a predefined and often narrow toolset fundamentally limits the model's ability to generalize. \textbf{(2) Programmatic manipulation}, where the model functions as a visual programmer, generating executable scripts for custom image manipulation~\citep{gupta2023visual, suris2023vipergpt,zhang2025thyme}. While theoretically Turing-complete, this paradigm is hindered by practical issues: high latency from the code generation-execution loop and poor expressiveness limited by pixel-level manipulation. %scalability due to the immense difficulty of curating diverse code-image training datasets. 
\textbf{(3) Generative imagination}, which leverages a model's internal generative capacity to synthesize new visual images as intermediate thoughts~\citep{li2025imagine,chern2025thinking, zhang2025latent}. However, this approach is not only computationally prohibitive but, more critically, risks ungrounded reasoning, as the synthesized content is generated from the model's internal priors rather than extracted from the source data. Our work, in contrast, offers a single, unified framework that is both scalable and expressive, leveraging language's compositional power to articulate cognitive intent beyond any fixed toolset, while simultaneously addressing cross-view information loss through dynamic, joint feature modulation.

% (e.g., Python code)
% Moreover, by directly manipulating feature maps, this strategy sidesteps the immense computational cost of full image synthesis, fostering a tightly integrated and efficient cognitive architecture. This charts a promising path toward a new class of AI that is not only more robust and flexible but also reasons \emph{with} vision in a truly profound way.

%% file: sec/3_method.tex
\section{Methodology}
\label{sec:methodology}
\begin{figure}[!t]
    \centering
    \includegraphics[width=0.95\linewidth]{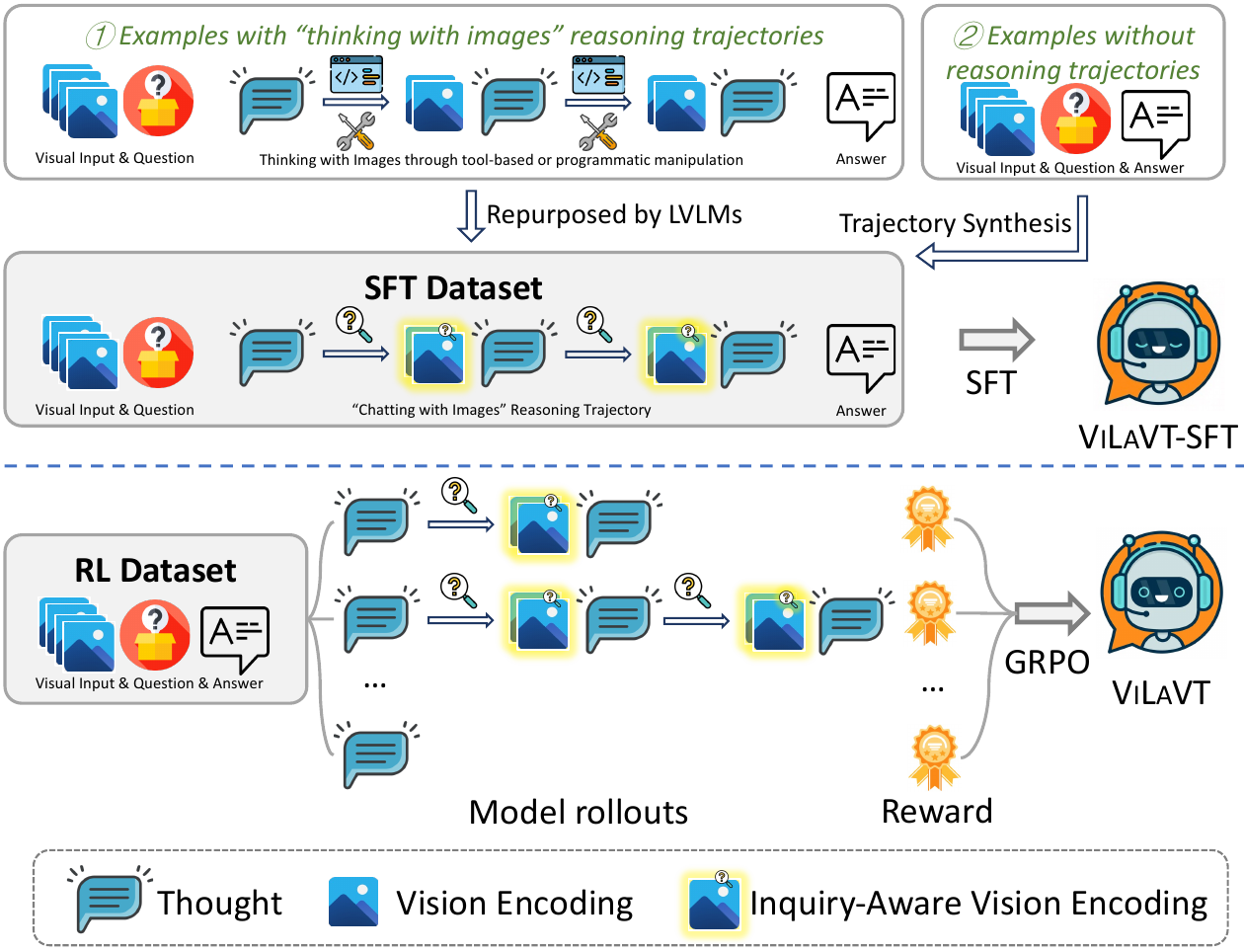}
    \caption{
    The two-stage training pipeline for \textsc{ViLaVT}, including supervised fine-tuning (SFT, \textbf{Top}), followed by reinforcement learning with the GRPO algorithm (RL, \textbf{Bottom}).}
    \label{fig:training}
\end{figure}

We formulate the visual reasoning task as follows: Given a question $Q$ and a visual input $\mathcal{I} = \{\mathcal{I}_n\}_{n=1}^N$ (where $N=1$ for a single image and $N>1$ for a video or image sequence), our goal is to generate the final answer $A$ by producing a reasoning trajectory $\tau$ interleaving textual thought and visual manipulation. To achieve this, we introduce the ``chatting with images'' reasoning paradigm (\S\ref{sec:paradigm}), which is underpinned by a novel {dynamic vision encoder} (\S\ref{sec:vit_arc}) and a {two-stage training strategy}, covering SFT (\S\ref{sec:sft}) and RL (\S\ref{sec:rl}). Figure~\ref{fig:hyperPara} and Figure~\ref{fig:training} illustrate the model architecture and the training pipeline, respectively.  %shows the overview of the training dataset. %%In contrast to the static single-image encoders of mainstream LVLMs~\citep{bai2025qwen2,hong2025glm}, our encoder is designed to interpret dynamically generated textual inquiries and jointly process multiple visual regions. 

\subsection{The ``Chatting with Images'' Paradigm}
\label{sec:paradigm}
Our framework unifies visual reasoning as a process of generating a trajectory $\tau = (s_1, s_2, \dots, s_T)$, where each step $s_t$ is a triplet $s_t = (r_t, q_t, z_t)$. The components are defined as:

\begin{itemize}[leftmargin=*]
    \item $r_t$: The model's textual reasoning process.
    \item $q_t$: A textual inquiry that directs the vision encoder to perform a targeted re-computation of visual features.
    \item $z_t$: A set of target visual regions, formally $z_t = \{(n_t^i, b_t^i)\}_{i=1}^{M_t}$. Here, $n_t^i$ is an index referencing a visual source, which can be an original image or any image produced in a preceding step. The term $b_t^i$ denotes a bounding box that specifies the sub-region to re-examine.
\end{itemize}
Following the generation of step $s_t$, the framework crops the set of visual regions $\mathcal{C}_t$ specified by $z_t$ from the preceding images. To enhance detail, each region is upscaled by a factor of 2 (capped at the original image dimensions) before being fed into the vision encoder $\mathcal{V}$. The encoder then jointly processes the textual inquiry $q_t$ and the upscaled regions $\mathcal{C}_t$ to produce a new feature set: $f_t = \mathcal{V}(\mathcal{C}_t, q_t)$. These resulting features $f_t$ are subsequently provided as input to the language model component, which we denote as $\mathcal{M}$, to inform the generation of the next step $s_{t+1}$. This iterative process is formally described as:
\begin{equation}
    s_{t+1} \sim \mathcal{M} \left(\cdot \mid f_0, Q, \{(s_k, f_k)\}_{k=1}^{t} \right)
\end{equation}
where $f_0 = \mathcal{V}(\mathcal{I}, \emptyset)$ represents the initial full-frame encoding, and the set $\{(s_k, f_k)\}_{k=1}^{t}$ constitutes the complete reasoning history up to the current step.

\subsection{The Architecture of the Vision Encoder}
\label{sec:vit_arc}
%is inherently static and processes a single image at a time. It 
%Our paradigm's visual manipulation hinges on a dynamic vision encoder to process text-conditioned, multi-region visual inputs jointly. 
A vanilla vision encoder~\citep{bai2025qwen2} processes each input image $\mathcal{I}_n$ independently. It first partitions the image into a sequence of patches, which are then linearly projected into patch embeddings $\boldsymbol{P}$. These embeddings are finally encoded by a vision Transformer~\cite{dosovitskiy2021an} to produce the visual representation.
% non-overlapping
% \in \mathbb{R}^{L \times D}

In contrast, the realization of our reasoning paradigm is predicated on a vision encoder with two key capabilities: (1) interpreting a textual inquiry $q$, and (2) jointly processing a variable number of potentially non-contiguous image regions. To fulfill these requirements, we re-architect the input stage of the vision encoder. Formally, given a set of $U$ images $\mathcal{C}=\{i_u\}_{u=1}^U$ and a textual inquiry $q$, the operation of our dynamic vision encoder $\mathcal{V}$ is formulated as:
\begin{align}
    \mathcal{V}(\mathcal{C}, q)=&\mathcal{V}_e(\boldsymbol{P}_1\oplus\boldsymbol{P}_2\oplus\cdots\boldsymbol{P}_U\oplus\boldsymbol{h}_q),\\
    \boldsymbol{h}_q =&\mathcal{V}_m(q),
\end{align}
where $\mathcal{V}_e$ is a vision Transformer and $\mathcal{V}_m$ is a lightweight language Transformer. Each image $i_u$ is projected into its patch embeddings $\boldsymbol{P}_u$. The inquiry $q$ is concurrently encoded by $\mathcal{V}_m$ into a query embedding $\boldsymbol{h}_q$. These embeddings are then concatenated %, interspersed with a special ``\texttt{[SEP]}'' token 
(denoted by $\oplus$), to form a unified input sequence. A left-to-right sequence of 2D positional embeddings~\cite{bai2025qwen2} is applied to this entire sequence, ensuring spatial and semantic coherence across all modalities. This elegant design enables the architecture to gracefully degenerate into a vanilla vision encoder when processing a single image without an inquiry ($U=1, q=\emptyset$). 
In contrast to InstructBLIP~\cite{li2023blip}, which uses a small set of learnable query tokens to query and compress dense visual features into a compact representation, our design enables query-guided cross-region interactions over multiple visual inputs during visual feature extraction.
% where $\mathcal{V}_e$/$\mathcal{V}_m$ is a vision Transformer and a lightweight language Transformer, respectively, each region $i_u$ is projected into a set of patch embeddings $\boldsymbol{P}_u$, the inquiry $q$ is encoded into textual embeddings $\boldsymbol{h}_q$ by $\mathcal{V}_m$, and the symbol $\oplus$ denotes concatenation through the embedding of a special token ``\texttt{[SEP]}''. The concatenated embedding sequence is then assigned a single, continuous set of two-dimensional positional embeddings, with indices increasing sequentially from left to right. This architecture gracefully degenerates to the vanilla vision encoder when processing a single image without an inquiry ($U=1, q=\emptyset$). 

Within the vision Transformer $\mathcal{V}_e$, we employ a hybrid attention strategy balancing computational cost and expressive power~\cite{bai2025qwen2}: in designated interaction layers, full self-attention is applied across all visual tokens, which also attend to the textual inquiry; in other layers, attention is restricted to operate either within each patch window independently or across the entirety of a single image, but not across distinct images.

% in several designated {interaction layers}, full self-attention is applied across all visual tokens while they also attend to the textual inquiry; in the remaining layers, attention reverts to the standard mechanism of processing each visual region independently for computational efficiency. 
%\textcolor{red}{(no separator tokens in current implementation)} 
% 其中，$\{i_u\}_{u=1}^U$表示$U$个图片，$\mathcal{V}_m$是一个小的语言模型用于编码模型推理产生的inquiry$q$，$\oplus$代表用一个特殊token的embedding拼接多图的patch embedding和文本的隐表示。当q是空，U=1时退化到standard ViT. 为了提升处理效率，Within the Transformer layers, we employ a directed attention mechanism: all visual patch embeddings 在做attention时能互相看到并更新， all visual patch embeddings can attend to but not update the textual representations $\mathcal{M}_v(q)$. 
%The 特殊分隔符 tokens are designed to be global information brokers; they attend to all visual and textual tokens and are attended to by all patch embeddings, enabling cross-region information exchange.

% \paragraph{Motivation.} This architectural design is central to our method's **expressiveness** and **efficiency**. By treating the vision encoder as a differentiable, on-demand information source conditioned by language, the model can perform arbitrary, fine-grained visual manipulations at the deep feature level. This fundamentally bypasses the pixel-level limitations and high latency of tool- or code-based approaches.
% You will need to include this package in your preamble:
% \usepackage{booktabs}

% You will need to include this package in your preamble:
% \usepackage{booktabs}
% \usepackage{multirow} % For spanning multiple rows

\subsection{Supervised Fine-tuning}
\label{sec:sft}
The goal of SFT is to bootstrap the model's ability to reason within the ``chatting with images'' paradigm by learning from expert demonstrations. To this end, we construct a large-scale, multi-domain dataset, $\mathcal{D}_{\text{SFT}}$, with reasoning trajectories from two distinct components.  Table~\ref{tab:dataset_stats} shows the overview of the training dataset. Appendix~\ref{data_construction} illustrates the data collection process in detail.

%operationalize this by
%This allows us to convert a wide array of such examples into our native format of ``chatting with images.'' Specifically, we 

The first component is constructed by repurposing existing ``thinking with images'' datasets featuring either tool-based or programmatic manipulation from the general domain. Since natural language serves as a universal interface capable of abstracting the functionality of any specialized tool or code block, we prompt a powerful teacher LVLM to translate each external action—be it a tool call~\citep{wang2025traceableevidenceenhancedvisual} or a code block~\citep{zhang2025thyme}—into an action triplet comprising the textual thought, a corresponding natural language inquiry, and the target visual regions. This process effectively unifies the heterogeneous landscape of interactive methods into our native ``chatting with images'' format.

% The second source of our SFT dataset specifically targets complex spatial reasoning tasks from datsets like~\citep{zhang2025flatland,ouyang2025spacerreinforcingmllmsvideo}. This domain serves as a crucial testbed, as static visual analysis often fails to capture the dynamic relationships that our ``Chatting with Images'' paradigm is designed to address (应该说这个任务更具挑战性，更需要模型通过visual thinking推导空间关系，不要说更适配我们的犯法). Consequently, a naive self-play approach for data generation is inadequate; a teacher model's own static vision encoder is incapable of producing the text-conditioned, multi-region interactions necessary to train our model's unique architecture（这里应该这么说：这些数据不提供reasoning trajectory，需要进行合成，以往的方法往往靠模型self-play，即用一个大模型rollout然后再进行outcome-based过滤，但我们没有这么做，原因是：xxx）.

The second SFT data component targets complex spatial reasoning datasets~\citep{zhang2025flatland,ouyang2025spacerreinforcingmllmsvideo,feng2025visuospatial}, which requires models to deduce intricate spatial and temporal relationships through deliberate visual thinking but lacks annotations of reasoning trajectories. The common practice to address this is rejection sampling~\cite{touvron2023llama2}, where a teacher LVLM generates entire reasoning trajectories which are then filtered based on the final outcome. However, we argue this method is sub-optimal, as a teacher model's static visual encoder cannot produce the text-conditioned, multi-region manipulations. Therefore, we design a dedicated pipeline that first programmatically mines latent spatial knowledge by performing parallel analyses (e.g., object grounding, camera motion estimation). This multi-faceted context is then synthesized into high-quality ``chatting with images'' trajectories, teaching our model to perform nuanced visual reasoning.

In addition, we augment the SFT dataset with purely textual trajectories generated via rejection sampling, teaching the model to answer simple questions efficiently when the visual information is self-evident~\citep{yang2025kwai}.
% The second source targets datasets that provide only question-answer pairs~\citep{zhang2025flatland,ouyang2025spacerreinforcingmllmsvideo}. To generate trajectories for these, we perform self-play using the teacher model with the ``chatting with image'' paradigm. A critical challenge here is that standard LVLMs' vision encoders do not support our text-conditioned, multi-region input. We thus have the teacher model's vision encoder process any newly generated image in a standard, non-conditional manner. We then apply strict filtering based on both format correctness and final answer accuracy to retain only high-quality, valid trajectories.
We then train the model, denoted as $\pi_\theta$, on this combined dataset via a standard maximum likelihood objective:
\begin{equation}
    \mathcal{L}_{\text{SFT}} = - {\mathbb{E}}_{(\mathcal{I}, Q, \{s_t\}_{t=1}^{T}) \sim \mathcal{D}_{\text{SFT}}} \sum_{t=1}^{T} \log \pi_\theta(s_t \mid \mathcal{I}, Q, s_{<t}).
\end{equation}
% \underset

\subsection{Reinforcement Learning}
\label{sec:rl}

%its purely imitative nature prevents it from exploring beyond the training data to discover better problem-solving strategies. 
%While SFT provides a strong foundation by learning from expert demonstrations, 
After SFT, we employ RL to explore the vast reasoning space and directly optimize for task outcomes.

\paragraph{Reward Function.}
%In the RL stage, %for a given visual input $\mathcal{I}$ and question $Q$, the model autoregressively generates a reasoning trajectory $\tau$. 
We evaluate each rollout trajectory using an outcome-based reward function: % $R(\tau)$: %, defined as:
\begin{equation}
    R(\tau) = \mathbbm{1}(R_{\text{correct}} > 0) (R_{\text{correct}} + R_{\text{format}}),
\label{eq:reward}
\end{equation}
First, the primary reward $R_{\text{correct}}$ measures the answer correctness, and the format reward $R_{\text{format}}$ evaluates the syntactic validity of the trajectory. The overall reward implements a strict gating mechanism based on the correctness of the final answer, which ensures the model cannot accumulate reward by generating syntactically perfect but ultimately incorrect reasoning paths~\citep{wu2025reinforcing}.

Specifically, the correctness reward, $R_{\text{correct}}$, is computed based on the task type, with a focus on tasks that have verifiable outcomes, including multiple-choice and numerical questions. For multiple-choice questions, $R_{\text{correct}}$ is a binary score (1/0) for an exact match. For numerical questions, we employ the Mean Relative Accuracy (MRA)~\citep{yang2024think} to provide a more granular signal. The format reward, $R_{\text{format}}$, is 1 if the entire trajectory is well-parsed and all bounding boxes are syntactically valid; otherwise, it is 0.
 % (e.g., normalized coordinates)

% To make this exploration tractable and avoid common failure modes in long-horizon reasoning, we introduce crucial early termination heuristics during policy rollouts. The rollout is immediately terminated with a zero reward if any of the following conditions are met:
% \begin{enumerate}
%     \item \textbf{Malformed Action:} The generated target region $z_t$ is syntactically or semantically invalid (e.g., coordinates outside $[0, 1]$ or defining a degenerate region).
%     \item \textbf{Excessive Length:} The trajectory length $T$ exceeds a predefined threshold $T_{\max}$.
%     \item \textbf{Redundancy:} The current action's inquiry-region pair duplicates a previous one, i.e., $(q_t, z_t) \in \{(q_k, z_k)\}_{k=1}^{t-1}$.
% \end{enumerate}

\paragraph{Optimization Object.} We optimize the policy $\pi_\theta$ using GRPO~\citep{shao2024deepseekmath}, a robust policy gradient algorithm without the KL penalty term~\citep{hu2025openreasonerzeroopensourceapproach}:
\begin{equation}
\begin{aligned}
    \mathcal{L}_{\text{RL}} &= -\mathbb{E}_{\tau \sim \pi_\theta} \left[ \sum_{t=1}^{T} \min\left(\rho_t A(\tau), \right.\right. \\ % 在 min 的逗号后换行
    & \qquad \qquad \qquad \left.\left. \text{clip}(\rho_t, 1-\epsilon_1, 1+\epsilon_2)A(\tau)\right) \right], \\
    \text{where} \quad \rho_t &= \frac{\pi_\theta(a_t|s_t)}{\pi_{\text{old}}(a_t|s_t)} \quad \text{and} \quad A(\tau) = \frac{R(\tau) - b}{\sigma + \delta}.
\end{aligned}
\label{rl_object}
\end{equation}
Here, $\rho_t$ is the importance sampling ratio, and $A(\tau)$ is the advantage computed by normalizing the trajectory reward $R(\tau)$ against the mean ($b$) and standard deviation ($\sigma$) of rewards from a batch of $G$ rollouts, with a small constant $\delta$ for numerical stability. %This normalization allows the model to effectively distinguish between high- and low-quality trajectories.

% \begin{equation}
%     \mathcal{L}_{\text{RL}}(\theta) = \mathbb{E}_{\tau \sim \pi_\theta} \left[ \sum_{t=1}^{T} \min(\rho_t A_t, \text{clip}(\rho_t, 1-\epsilon, 1+\epsilon)A_t) \right]
% \end{equation}
% where $\rho_t = \frac{\pi_\theta(a_t|s_t)}{\pi_{\theta_{\text{old}}}(a_t|s_t)}$ is the importance sampling ratio, and $A_t$ is the normalized advantage estimate.

%% file: sec/4_experiment.tex
\section{Experiments}

\subsection{Experimental Setup}
\begin{table*}[t!]
    \centering
    \caption{
        Main results comparing \textsc{ViLaVT} against baselines across a comprehensive suite of benchmarks. We report accuracy (\%) and highlight the best performance in \textbf{bold}; the second-best is \underline{underlined}.
    }
    \label{tab:main_results}
    % \sisetup{table-format=2.1, table-column-width=1.1cm}
    
    \begin{adjustbox}{width=\textwidth}
    
    % 定义12列
    \begin{tabular}{l cccccccc}
        \toprule[1.5pt]
        \multicolumn{1}{c}{\multirow{3}{*}{\textbf{Model}}} & \multicolumn{2}{c}{\textbf{General VQA}} & \multicolumn{6}{c}{\textbf{Spatial Reasoning}} \\%& \multirow{3}{*}{\textbf{Avg.}} \\
        \cmidrule(lr){2-3} \cmidrule(lr){4-9}
        
          & \multicolumn{2}{c}{\textbf{Single-Image}}& \multicolumn{2}{c}{\textbf{Single-Image}} & 
          \multicolumn{3}{c}{\textbf{Multi-Image}} & \multicolumn{1}{c}{\textbf{Video}} \\
        \cmidrule(lr){2-3} \cmidrule(lr){4-5} \cmidrule(lr){6-8}\cmidrule(lr){9-9}

         & HRBench-4K & HRBench-8K & {SpatialEval-Real} & EmbSpatial & ERQA  & SPAR-Bench & MMSI-Bench & 
         % ViewSpatial & 
         VSI-Bench\\
        \midrule[1.5pt] 
        % \midrule
        % \midrule
        \multicolumn{9}{c}{\textbf{Non-Thinking}}\\
        \midrule

         Qwen2.5-VL-7B & 67.8 & 65.1 & 58.5 & 52.7 & 39.3  & 36.9 & 26.9  &  34.7 \\
         InternVL3-8B & 70.8 & 62.0 & 62.3 & 66.9 & 35.3  & 36.0 & 25.7 &  42.1  \\
         LLaVA-OneVision-7B & 64.3 & 59.8 & 62.9 & \textbf{72.5}  & 30.6 & 32.4 & 24.5 &  32.4  \\
        \midrule[1.5pt] 
        
        \multicolumn{9}{c}{\textbf{Thinking about Images}}\\
        \midrule
         Qwen2.5-VL-7B & 69.8 & 64.9 & 54.1 & 58.2 & 39.8 & 31.6 & 27.1  &  26.2 \\
        SpaceR-7B & 58.1 & 49.8 & 62.7 & \underline{69.4}  & 40.3 & 37.1 & 28.8 & 45.6 \\
        Spatial-MLLM-4B & - & - & - & -  & 38.3 & 35.1 & 27.7 &  \underline{48.4} \\
        
        \midrule[1.5pt] 
        \multicolumn{9}{c}{\textbf{Thinking with Images}}\\
        \midrule
         {\textsc{ViLaSR}-7B}  & 60.5 & 56.3 & \underline{63.9} & 66.4 & 41.0 & 37.6 &  \underline{30.2}  & 45.4 \\
        
        % Baseline Group 3: Programmatic
        Pixel-Reasoner-7B & 72.9 & 66.9 & 61.6 & 62.3 & - & - & - & - \\
        DeepEyes-7B & 75.1 & \textbf{72.6 }& 62.8 & 64.2 & - & - & - & -  \\
        Thyme-7B  & \textbf{78.3} & \underline{72.4} & 63.7 & 65.5 & - & - & - & -   \\
        % \midrule[1.5pt] 
        % \multicolumn{10}{c}{\textbf{Specialized Spatial Model}}\\
        % \midrule
        \midrule[1.5pt] 
        \multicolumn{9}{c}{\textbf{Chatting with Images (Ours)}}\\
        \midrule
        % % Your Model
        % {\textsc{ViLaVT}-7B} & \underline{75.5} & 69.3 & \textbf{64.4} & 67.3 & \textbf{42.2} & \textbf{45.4} & \underline{31.3} & \textbf{51.3}  \\
        {\textsc{ViLaVT}-7B} & \underline{75.5} & 69.3 & \textbf{68.9} & 69.3 & \textbf{42.2} & \textbf{52.6} & \textbf{31.3} & \textbf{52.0}  \\
        % {\textsc{ViLaVT}-SFT} & 73.6 & 66.2 & 61.5 & 66.2 & 40.8 &   & 45.4 & 28.6 & 40.5& 43.3 \\
        \bottomrule[1.5pt]
    \end{tabular}
    \end{adjustbox}
\end{table*}

\paragraph{Evaluation Benchmarks and Metrics.}
To comprehensively evaluate our model's capabilities, we select a diverse suite of benchmarks spanning different reasoning complexities and task formats:
\begin{itemize}[leftmargin=10pt]
    \item \textbf{General visual question-answering (VQA):} This category of tasks assesses the fundamental ability for real-world fine-grained perception and understanding, including \textit{HRBench-4K} and \textit{HRBench-8K}~\citep{wang2025divide}. %\textbf{MME-RealWorld} series~\citep{zhang2025mmerealworld} \textbf{HRBench}~\citep{wang2025divide}, \textbf{V*}~\citep{wu2024v}, and \textbf{RealWorld QA}~\citep{realworldqa2024xai}.

    % \item \textbf{Mathematical and Logic Reasoning:} This category evaluates the model's capacity for abstract reasoning grounded in visual context. We use \textbf{MathVision}~\citep{wang2024mathvision}, \textbf{MathVista}~\citep{lu2024mathvista}, and \textbf{MathVerse}~\citep{zhang2024mathverse}, which contain visually-rich mathematical problems; and \textbf{LogicVista}~\citep{xiao2024logicvista}, \textbf{WeMath}~\citep{qiao2024wemathdoeslargemultimodal} and \textbf{VisuLogic}~\citep{xu2025visulogic} to specifically probe formal logical reasoning abilities.

    % \item \textbf{Spatial reasoning:} This suite, comprising \textbf{VSI-Bench}~\cite{yang2024think}, \textbf{ERQA}~\citep{team2025gemini}, and \textbf{MindCube}~\citep{yin2025mindcube}, specifically probes the model's understanding of complex spatial relationships across single images, multiple images, and videos.
% \textit{MindCube}~\citep{yin2025mindcube}
    \item \textbf{Spatial Reasoning:} This suite is designed to probe the model's understanding of complex spatial relationships across diverse visual modalities. It is organized into three categories: 
(i) \textbf{single-image} reasoning: \textit{SpatialEval-Real}~\citep{wang2024isapicture} and \textit{EmbSpatial}~\citep{du-etal-2024-embspatial}; 
(ii) \textbf{multi-image} reasoning: \textit{ERQA}~\citep{team2025gemini}, \textit{SPAR-Bench}~\citep{zhang2025flatland}, and \textit{MMSI-Bench}~\citep{yang2025mmsibenchbenchmarkmultiimagespatial}; 
and (iii) \textbf{video-based} reasoning: \textit{VSI-Bench}~\citep{yang2024think}.
    % \item \textbf{GUI Grounding:} It requires precise grounding of natural language commands to specific UI elements on a screen, involving ScreenSpot-v2~\cite{} and ScreenSpot-Pro~\citep{}.
\end{itemize}
For evaluation metrics, we report \textbf{Accuracy (Acc.)} and \textbf{Mean Relative Accuracy (MRA)}~\citep{yang2024think} for multiple-choice and numerical questions, respectively. Appendix~\ref{benchmark_details} shows the benchmark statistics.

\paragraph{Implementation Details.}
Our model architecture is initialized from Qwen2.5-VL-7B~\citep{bai2025qwen2}. The lightweight language Transformer $\mathcal{V}_m$ in our vision encoder is initialized from Qwen3-0.6B-Embedding~\citep{yang2025qwen3} and kept frozen during training. During the training phase, we allocate a total budget of 8,192 visual tokens across the initial visual inputs to set each image's maximum resolution, which is then kept fixed throughout multi-turn reasoning; for video tasks, we uniformly sample 32 frames per clip. 
For supervised training stage, we train the model for 2 epochs with a learning late of $1\times10^{-5}$ and  a global batch size of 384.  The subsequent
RL optimization is implemented using the VERL framework~\citep{sheng2024hybridflow}, where we set the rollout batch size to 96 and generate 4 candidate reasoning paths per query. Additional details are provided in the Appendix~\ref{implementation_details}. We also present the training dynamics in Section~\ref{sec:training_dynamics}.

\paragraph{Baselines.}
We compare \textsc{ViLaVT} against a comprehensive set of baselines spanning distinct reasoning paradigms:
\begin{itemize}[leftmargin=10pt, topsep=2pt, itemsep=0pt]
    \item \textbf{Non-thinking:} The models generate the final answer in a single step without producing an explicit reasoning chain. We include the powerful open-source LVLMs, {Qwen2.5-VL}~\citep{bai2025qwen2}, InternVL3~\citep{zhu2025internvl3}, and LLaVA-OneVision~\citep{li2024llava}.
    \item \textbf{Thinking about images:} The models perform textual reasoning to deduce the final answer based on a one-pass, static visual encoding, including {Qwen2.5-VL}~\citep{bai2025qwen2} and two specialized models for spatial reasoning, i.e., SpaceR~\cite{ouyang2025spacerreinforcingmllmsvideo} and Spatial-MLLM~\cite{wu2025spatial}.
    \item \textbf{Thinking with images:} 
    % These models dynamically manipulate the visual input during the reasoning process. 
    We evaluate against four representative approaches: {\textsc{ViLaSR}}~\citep{wu2025reinforcing},  Pixel-Reasoner~\citep{su2025pixel} and DeepEyes~\citep{zheng2025deepeyes}, which leverage external tools for visual manipulation,  and {Thyme}~\citep{zhang2025thyme}, which generates Python code for programmatic manipulation. 
\end{itemize}
% In addition, appendix~\ref{ablation_study} compares our full model with its ablated versions. %{\textsc{ViLaVT}-SFT}, which is trained only with SFT,  To isolate the contributions of RL, .

\subsection{Main Results}
\label{sec:main_results}
% (1) \textbf{The ``Thinking about Images'' paradigm offers negligible, and at times detrimental, performance changes.} Simply prompting a model to ``think'' with text only is insufficient to overcome the limitations of its static visual encoding. For instance, the performance of Qwen2.5-VL-7B only marginally changes on HRBench-4K (67.8\% to 69.8\%), while it significantly degrades on key spatial tasks like VSI-Bench (34.7\% to 26.2\%).
Table~\ref{tab:main_results} shows the main results, highlighting three clear conclusions:
% \textbf{(1) ``Thinking with images'' methods show task-specific specialization and lack universality.} While these methods achieve better average performance than ``thinking about images'' baselines, no single approach excels across the board. For example, Thyme-7B leads on General VQA with 78.3\% on HRBench-4K but is not the top performer on spatial tasks. ViLaSR-7B is strong on spatial benchmarks but lags significantly on general VQA (60.5\% on HRBench-4K). This highlights a clear trade-off between models optimized for either fine-grained or high-level reasoning.
% \textbf{(2) ``Chatting with images'' achieves superior overall performance.} \textsc{ViLaVT} consistently establishes a new state-of-the-art across all benchmarks. By combining the high expressiveness of language prompting with the relational power of joint multi-image encoding, it surpasses specialized models even in their own domains.
\textbf{(1) Prior paradigms exhibit task-specific specialization.} Models in both ``thinking about images'' and ``thinking with images'' paradigms often excel in narrow domains. For example, Thyme and DeepEyes lead on HRBench but perform weakly on complex spatial reasoning. Conversely, SpaceR and Spatial-MLLM, which are specialized for spatial tasks, lag on general VQA. Such specialization often stems from a lack of a unified and expressive interface for visual interaction, forcing model designs to be narrowly tailored to specific task structures.
\textbf{(2) ``Chatting with images'' achieves superior versatility and overall performance.} \textsc{ViLaVT} demonstrates remarkable versatility. It establishes new state-of-the-art results on 5 of 8 benchmarks, particularly in complex multi-image and video-based spatial reasoning. While highly specialized models like Thyme may achieve slightly higher scores on certain single-image VQA tasks, \textsc{ViLaVT} remains highly competitive. %This strong, consistent performance across a diverse range of domains underscores the power of our paradigm. 
By unifying visual interaction within a language-guided framework, \textsc{ViLaVT} emerges as a more powerful and general-purpose visual reasoner.
% These results provide compelling evidence that our paradigm offers a more powerful and versatile path toward advanced visual reasoning. 
% ablation studies in Appendix~\ref{ablation_study} validate the distinct contributions of our key components. 
Additionally, Appendix~\ref{overhead_analysis} provides an analysis of the computational overhead introduced by these reasoning paradigms.

\begin{table}[]
    \centering
    \caption{Ablation study on training stages and model components. 
    % Our full model (ViLaVT) consistently outperforms variants.
    }
    \label{tab:ablation}
    \resizebox{\linewidth}{!}{
    \begin{tabular}{lcccc}
        \toprule
        \textbf{Model} & \textbf{HRBench-4K} & \textbf{HRBench-8K} & \textbf{ERQA} & \textbf{VSI-Bench} \\
        \midrule
        \textbf{Qwen2.5-VL-7B}        & 67.8 & 65.1 & 39.3 & 34.7 \\
        \textbf{Qwen2.5-VL-7B-SFT}    & 68.8 & 63.6 & 38.8 & 46.9 \\
        \textbf{\textsc{ViLaVT}-SFT}  & 73.6 & 66.1 & 39.8 & 43.3 \\
        \textbf{\textsc{ViLaVT}}      & \textbf{75.5} & \textbf{69.3} & \textbf{42.2} & \textbf{52.0} \\
        \midrule
        \textbf{\textsc{ViLaVT}$_{400}$ w/ vanilla ViT} & 73.3 & 65.7 & 40.3 & 47.4 \\
        \textbf{\textsc{ViLaVT}}$_{400}$ & 73.4 & 67.0  & 40.7 & 49.0  \\
        \bottomrule
    \end{tabular}}
\end{table}

\begin{figure}[t]
    \centering
    \includegraphics[width=0.93\columnwidth]{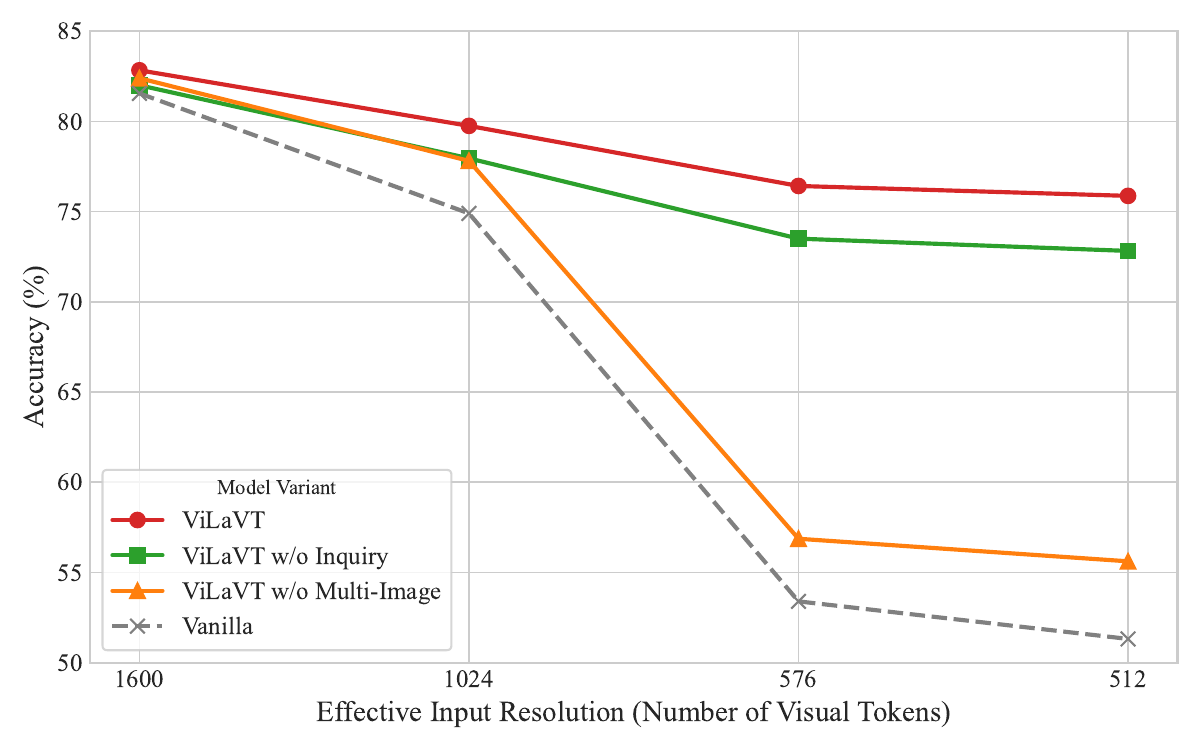}
    \caption{Vision encoder analysis across resolutions. Our full model exhibits increasingly performance gains over ablations as resolutions decreases, evidencing robustness to information loss.}
    % \caption{
    %     Vision encoder analysis under varying input resolutions. Our full model demonstrates increasing performance gains over the ablated models as resolution decreases, highlighting its robustness to information loss.} %Both inquiry-conditioning and multi-image capabilities prove to be essential.
    % }
    \label{fig:ablation_vision_encoder}
\end{figure}

\subsection{Ablation Studies on Training Stages and Model Components}

To validate our design choices, we conduct ablation studies on five variants:
(1) \textbf{Qwen2.5-VL-7B}: the base model; 
(2) \textbf{Qwen2.5-VL-7B-SFT}: the base model SFT-trained on the same corpus using only question–answer pairs with no reasoning chains;
(3) \textbf{\textsc{ViLaVT}-SFT}: our model trained with SFT only; 
Additionally, to isolate the contribution of our dynamic vision encoder, we train two variants with the same full SFT stage and 400-step RL schedule, in contrast to full 1200-step RL stage for \textsc{ViLaVT}:
(4) \textbf{\textsc{ViLaVT}$_{400}$ w/ vanilla ViT}, which replaces our dynamic encoder with a standard ViT, and 
(5) \textbf{\textsc{ViLaVT}}$_{400}$, which retains the full model architecture but trains the RL stage for only 400 steps.
The results in Table~\ref{tab:ablation} reveal several key insights:

\noindent\textbf{Importance of our paradigm.} 
Comparing Qwen2.5-VL-7B-SFT with \textsc{ViLaVT} trained on the same corpus, Qwen2.5-VL-7B-SFT yields only a modest $1.0\%$ on HRBench-4K and even drops on HRBench-8K relative to the base model. In contrast, \textsc{ViLaVT} consistently outperforms Qwen2.5-VL-7B-SFT, showing that the gains mainly stem from our framework rather than data scaling alone.
% To isolate the contribution of our method from that of the data, we compare Qwen2.5-VL-7B-SFT with \textsc{ViLaVT}. Although Qwen2.5-VL-7B-SFT is trained on the same fine-tuning and RL data, it yields only marginal or even negative changes over the base model on several benchmarks (e.g., a modest +1.0 improvement on HRBench-4K and a performance drop on HRBench-8K). In contrast, our \textsc{ViLaVT} delivers clear and consistent gains over Qwen2.5-VL-7B-SFT. This demonstrates that simply adding more data is insufficient; the main improvements stem from our framework—specifically, the architectural modifications, the reasoning paradigm, and the proposed training pipeline—which are crucial for effective  reasoning.

\noindent\textbf{Importance of training stages.}  \textsc{ViLaVT}-SFT achieves substantial improvements over the base model, indicating that supervised fine-tuning effectively enables the model to acquire the proposed interactive format. Building upon this foundation, \textsc{ViLaVT} further improves over \textsc{ViLaVT}-SFT across all benchmarks, suggesting that RL refines the model’s interactive reasoning beyond supervised learning.

\noindent\textbf{Importance of our vision encoder.}
Under the same full SFT + 400-step RL training schedule, we observe that \textsc{ViLaVT}$_{400}$ consistently outperforms \textsc{ViLaVT}$_{400}$ w/ vanilla ViT, especially on HRBench-8K and VSI-Bench, indicating that the gains come not only from iterative reasoning but also from query-conditioned joint encoding within our vision encoder.

\subsection{Analysis under Different Resolutions}
\label{sec:encoder_analysis}

\begin{table}[t]
    \centering
    \caption{Ablations on \textit{HRBench-4K} across image resolutions.}
    \label{tab:hrbench4k}
    \resizebox{\linewidth}{!}{
    \begin{tabular}{c c c c c }
        \toprule
        \textbf{Input Resolution} & \textbf{Qwen2.5-VL-7B} & \textbf{\textsc{ViLaVT}-SFT}  &  \textbf{\textsc{ViLaVT}} \\
        \midrule
        1024 & 56.4 & 62.5 & \textbf{63.9} \\ 
        2048 & 61.4 & 69.8  & \textbf{70.1} \\ 
        4096 & 66.8 & 69.9   & \textbf{73.0} \\ 
        8192 & 67.8  & 73.6   & \textbf{75.5} \\ 
        \bottomrule
    \end{tabular}}
\end{table}

To isolate and validate the effectiveness of our dynamic vision encoder, we first conduct a controlled experiment on a subset of the SPAR dataset~\cite{zhang2025flatland}, comprising 6,489 training and 721 testing examples, each with 2-4 associated images. All models are trained on this subset for 5 epochs to directly predict the final answer, bypassing any textual reasoning chain. The training for all models used a fixed maximum input resolution of $1600 \times 28 \times 28$ pixels, corresponding to 1600 visual tokens.
We compare \textsc{ViLaVT} with three ablations: (1) \textbf{\textsc{ViLaVT} w/o Inquiry}, removing inquiry-conditioning; (2) \textbf{\textsc{ViLaVT} w/o Multi-Image}, removing multi-image interaction; and (3) \textbf{Vanilla}, removing both. 
As shown in Figure~\ref{fig:ablation_vision_encoder}, performance drops as resolution decreases for all models, but \textsc{ViLaVT} remains substantially more robust; at 512 tokens, it achieves 75.9\% accuracy creating a {24.8\%} performance gap over the vanilla model's {51.1\%}. The ablations further suggest that joint multi-image encoding is the main driver of robustness on this multi-view benchmark, while inquiry-conditioning provides a consistent additional gain.
These results that our architectural modifications are critical for preserving performance in information-scarce scenarios\footnote{Although effective for static multi-view inputs, this single-pass encoding is computationally prohibitive for long videos and structurally unsuited for multi-hop reasoning. 
Therefore, our main experiments employ the full iterative paradigm, which starts with a vanilla visual encoding and then performs dynamic, targeted re-encoding during reasoning to balance performance with scalability.
}.

% \footnote{Although powerful, this single-pass encoding approach faces practical limitations: it is computationally prohibitive for long videos and is structurally unsuited for multi-hop reasoning. Therefore, our main experiments employ the full iterative paradigm, which starts with a vanilla visual encoding and then performs dynamic, targeted re-encoding during reasoning to balance performance with scalability.}

To validate the generalizability of these findings, we further evaluate \textsc{ViLaVT} on \textit{HRBench-4K} under varying token budgets (Table~\ref{tab:hrbench4k}). \textsc{ViLaVT} consistently outperforms the base Qwen2.5-VL-7B across all resolutions, including information-scarce settings (e.g., 63.9\% vs.\ 56.4\% at 1,024 tokens). Moreover, the performance gap between \textsc{ViLaVT}-SFT and the full \textsc{ViLaVT} persists across resolutions, demonstrating that RL refines the model's reasoning strategy under different resolution constraints.

\begin{figure}
    \centering
    \includegraphics[width=0.85\columnwidth]{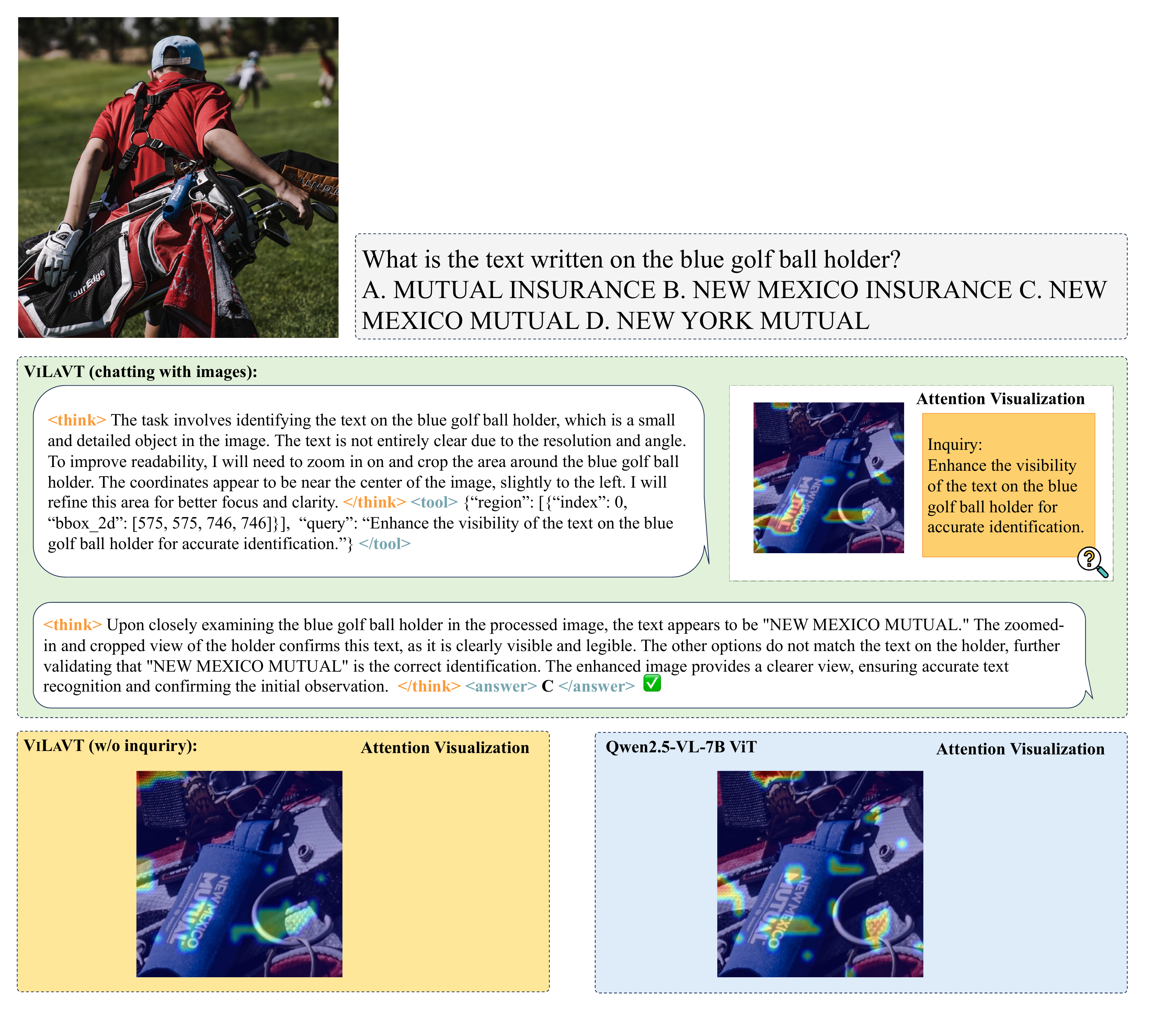}
    \caption{Attention visualization on an HRBench-4K example.}
    \label{fig:attn_case_19}
\end{figure}

\subsection{Case Study}

\subsubsection{Representative Case}

Figure~\ref{fig:case_study} qualitatively demonstrates our paradigm's strength on a complex video-based spatial reasoning task. \textsc{ViLaVT} successfully solves the problem by using highly expressive language prompts to express clear cognitive intent, which then guide a joint, relational re-encoding of features across key frames (e.g., comparing Images 25-27 against Image 5). This enables the model to build a coherent spatial map. We provide further comparisons with other ``thinking with images'' methods in Appendix~\ref{app:more_cases}.
% illustrating how \textsc{ViLaVT} bridges high-level declarative intent, articulated through language prompts, with in-depth relational analysis, facilitated by joint feature re-encoding.

\subsubsection{Attention Map Visualization}
Figure~\ref{fig:attn_case_19} illustrates that inquiry-conditioned encoding syields sharper attention on task-relevant regions. Conditioned on the textual inquiry, \textsc{ViLaVT} progressively focuses on task-relevant regions---first localizing the blue golf ball holder, then focus on its embedded text during re-encoding. In contrast, \textsc{ViLaVT} w/o Inquiry and Qwen2.5-VL-7B with a vanilla ViT backbone exhibit more diffuse attention. This qualitative evidence supports that conditioning the vision encoder on the inquiry improves localized perception and facilitates fine-grained detail extraction. Additional cases and visualization details are provided in Appendix~\ref{app:visualization_cases}.

% We also visualize the attention maps of the \textsc{ViLaVT} vision encoder to better understand how our framework improves query-aware perception. Concretely, we extract attention maps from the last layer of the vision encoder, average them across all heads and all image patches, and overlay the resulting heatmap on the original image.

% In Figure~\ref{fig:attn_case_19}, \textsc{ViLaVT} first selects the region containing the blue golf ball holder, which includes potentially relevant text, and chooses it as the area to zoom in. In the second round, when re-encoding the zoomed-in image, the query-aware vision encoder further concentrates its attention on the text area, enabling the model to capture fine-grained textual details that are crucial for answering the query.
% By contrast, \textsc{ViLaVT} w/o Inquiry and Qwen2.5-VL-7B with the vanilla ViT backbone show weaker or more dispersed attention over the text region, indicating less effective focus on query-relevant content. These qualitative patterns highlight the benefit of our architectural modifications: injecting the textual query into the vision encoder yields sharper, more localized attention on task-critical regions and facilitates fine-grained detail extraction.
% Additional attention visualization cases are provided in Appendix~\ref{app:visualization_cases}.

%% file: sec/5_conclusion.tex
\section{Conclusion}

In this work, we introduce ``chatting with images,'' a unified and scalable visual reasoning paradigm designed to overcome the dual limitations of modern approaches: the information loss in static ``thinking about images'' and the low expressiveness of visual prompts within ``thinking with images''. We reframe visual manipulation as language-guided feature modulation, bridging high-level declarative intent, articulated through expressive language prompts, with deep relational analysis enabled by joint feature re-encoding. Comprehensive experiments validate the superiority of our approach, which establishes new state-of-the-art results on most evaluated benchmarks. Looking ahead, we believe this paradigm opens promising avenues for developing more compositional and general-purpose visual reasoning agents.

%% file: sec/6_suppl.tex
% \section{Appendix}

\clearpage
\appendix
% \setcounter{page}{1}
% \maketitlesupplementary
% \section{Appendix}
\section{Data Construction}\label{data_construction}

\begin{table*}[h!]
\centering
\caption{Statistics of the dataset for our two-stage (SFT+RL) training. 
The SFT stage utilizes two types of reasoning trajectories: standard textual-only and our proposed ``Chatting with Images.'' We either repurpose existing trajectories from source datasets (\ding{51}) or synthesize both types from scratch where none are provided (\ding{55}). \textbf{MC:} Multiple-Choice.}
\label{tab:dataset_stats}
\begin{adjustbox}{width=\textwidth} 
\begin{tabular}{@{}lccccccc@{}}
\toprule
\multirow{2}{*}{\textbf{Source}} & \multirow{2}{*}{\textbf{Domain}}&\multirow{2}{*}{\textbf{Visual Input}} & \multirow{2}{*}{\textbf{Task Type}}&{\textbf{Trajectory}} & \multicolumn{2}{c}{\textbf{SFT}} & \multirow{2}{*}{\textbf{RL}} \\
\cmidrule(lr){6-7} %\cmidrule(l){8-8}
 & & & &\textbf{Provided}& \textbf{Textual} & \textbf{Chatting with Images} &  \\
\midrule
\textbf{{VGR}}~\citep{wang2025vgr} & General & Single Image & MC & \ding{51} & 0 & 30,090 & 8,629 \\
\textbf{{Thyme}}~\citep{zhang2025thyme} & General & Single Image & MC& \ding{51}  & 96,402 & 70,375 & 9,378 \\
\midrule
\textbf{{SpaceR}}~\citep{ouyang2025spacerreinforcingmllmsvideo} & Spatial Reasoning & Video & MC / Numerical& \ding{55}  & 11,629 & 25,973 & 19,316  \\
\textbf{{SPAR}}~\citep{zhang2025flatland} & Spatial Reasoning & Video & MC / Numerical& \ding{55}   & 9,280 & 26,607 & 13,779 \\
\textbf{{Vica}}~\citep{feng2025visuospatial} & Spatial Reasoning & Video & MC / Numerical& \ding{55}   & 25,307 & 11,332 & 68,898 \\
\bottomrule
\end{tabular}
\end{adjustbox}
\end{table*}

Our Supervised Fine-Tuning (SFT) dataset, $\mathcal{D}_{\text{SFT}}$, is meticulously constructed from three distinct sources to provide our model with a comprehensive and diverse set of reasoning skills. This section details the construction process for each component.

\subsection{Component 1: Repurposing ``Thinking with Images'' Trajectories}
The first component is constructed by repurposing existing ``thinking with images'' datasets from the general domain, which feature either tool-based~\citep{wang2025traceableevidenceenhancedvisual} or programmatic manipulation~\citep{zhang2025thyme}. The core idea is to translate the procedural, external actions from these datasets into our declarative, internal ``chatting with images'' format. We leverage the hypothesis that natural language serves as a universal interface capable of abstracting the functionality of any specialized tool or code block.

\begin{figure*}
    \centering
    \includegraphics[width=\textwidth]{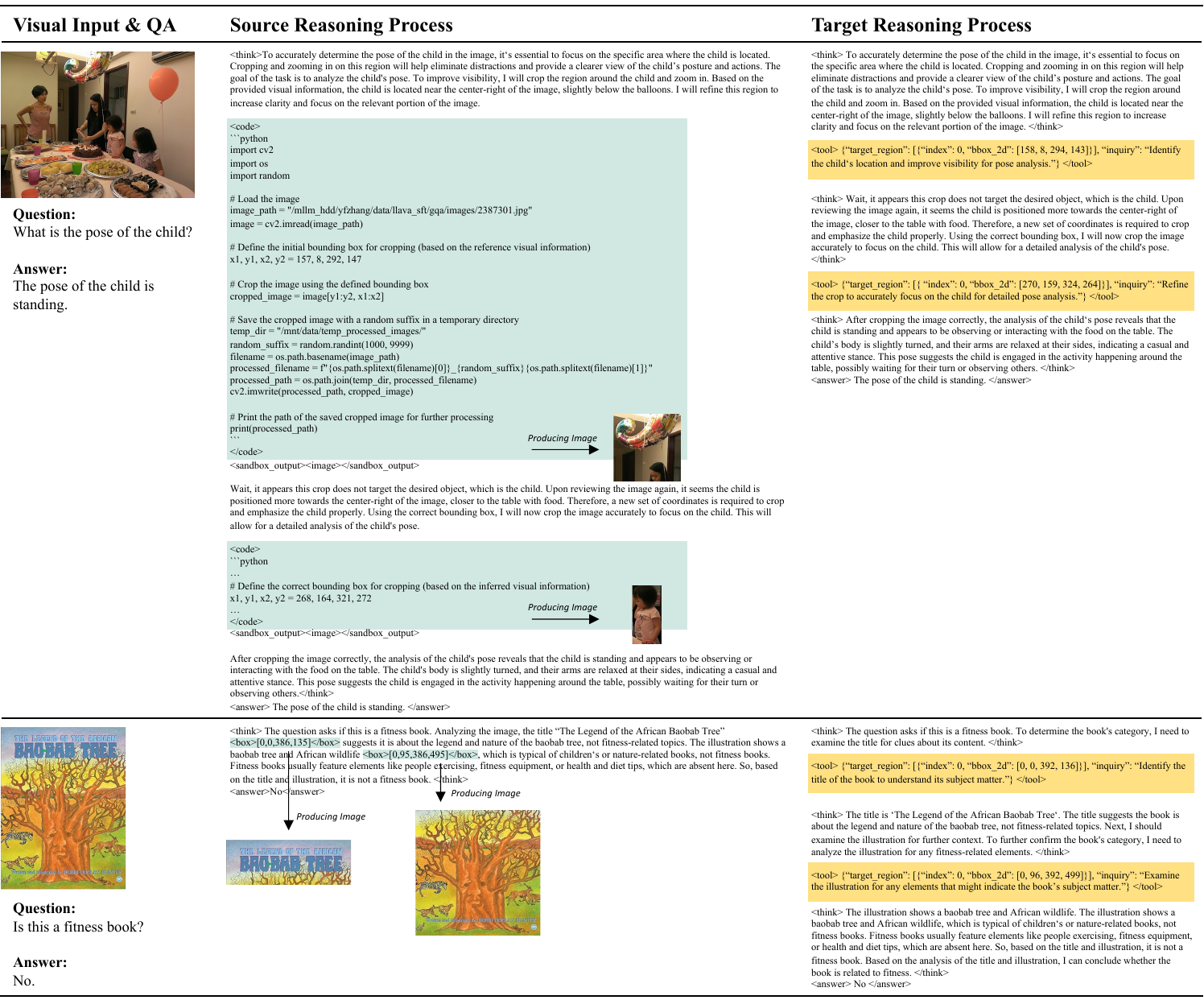}
    \caption{
{Illustration of our data unification process for SFT.}
This figure showcases how we convert heterogeneous, externally-interactive reasoning trajectories from existing datasets into our native ``Chatting with Images'' format. 
For each step in the \texttt{Source Reasoning Process} that involves an external action (in light green), we prompt a teacher model to translate it into our internal action format (in light orange). 
This target format consists of a natural language \texttt{inquiry} that captures the {intent} of the original action, and the corresponding \texttt{target\_region} data.
\textbf{Top Example} demonstrates the conversion of a complex, multi-step Python sandbox session. Note how the process handles a self-correction loop (the first crop is incorrect, leading to a second, refined action), translating procedural code execution into a sequence of declarative, language-guided inquiries.
\textbf{Bottom Example} shows the same process applied to a simpler, tag-based tool, demonstrating the {versatility} of our unification strategy across different source formats.
The highlighted sections in the \texttt{Target Reasoning Process} denote the generated \texttt{<tool>} calls, which form the core training signal for our model.
}
    \label{fig:convert_example}
\end{figure*}

To achieve this, we prompt a powerful teacher LVLM, Qwen2.5-VL-72B~\cite{bai2025qwen2}, to translate each external action (i.e., either a tool call or a code block) into an action triplet comprising (1) the textual thought, (2) a corresponding natural language inquiry, and (3) the target visual regions. The prompt used for this conversion is detailed in Figure~\ref{fig:full_translation_prompt}. This process effectively unifies the heterogeneous landscape of interactive methods into our native format. Figure~\ref{fig:convert_example} provides concrete examples of this conversion process. %, showing how a complex Python session and a simple tool call are both seamlessly converted into our model's native reasoning steps.

% A figure for the prompt. I'm using `tcolorbox` for a nice framed box.
% Make sure you have `\usepackage[most]{tcolorbox}` in your preamble.
% 在导言区确保已经加载了 \usepackage[most]{tcolorbox}

\begin{figure*}[t!]
    \centering
    % System Prompt Box
    % \begin{tcolorbox}[
    %     colback=blue!5,
    %     colframe=blue!75!black,
    %     width=\textwidth,
    %     arc=2mm,
    %     auto outer arc,
    %     title=\textbf{System Prompt: Guiding Principles for ``Thinking with Images'' Data Conversion},
    %     fonttitle=\bfseries,
    %     boxsep=2mm,
    %     toptitle=2mm,
    %     bottomtitle=2mm,
    %     left=4mm
    % ]
    % \small
    % \texttt{You are a highly intelligent data extraction and reasoning agent. Your task is to analyze a conversation between a user and a tool/code-using AI model and convert it into a structured JSON format. You must adhere to the following principles:}
    % \begin{enumerate}
    %     \item \textbf{Semantic Action Inference:} \texttt{For each code block or bounding box, your primary goal is to infer its {intent}, not just its literal function. Crucially, this field must describe the question the model is trying to answer with the code/box, not the answer or observation it finds.}
    %     \item \textbf{Strict JSON Adherence:} \texttt{Your final output must be a single, valid JSON object... Do not include any explanations...}
    % \end{enumerate}
    % \end{tcolorbox}
    
    % \vspace{2mm} % Add some vertical space between the boxes

    % User Prompt Box
    \begin{tcolorbox}[
        colback=green!5,
        colframe=green!60!black,
        width=\textwidth,
        arc=2mm,
        auto outer arc,
        title=\textbf{Prompt for ``Thinking with Images'' Data Conversion},
        before upper={\small},  % <--- 将字体命令作为选项
        fonttitle=\bfseries,
        boxsep=1mm,
        toptitle=1mm,
        bottomtitle=1mm,
        left=4mm
    ]
    \scriptsize
    \texttt{You are a highly intelligent data extraction agent. Please analyze a conversation between a user and a tool- or code-using AI model, and convert it into a structured JSON format. Given the user instruction and model response below, extract a list where each element corresponds to one `<code>` block or `bounding box`.}

    \textbf{Field Definitions:}
    \begin{itemize}
        \item \texttt{\textbf{thought}: The high-level reasoning or planning text that precedes a code/box. It describes the model's internal monologue, connecting previous observations to the motivation for the upcoming inquiry.}
        \item \texttt{\textbf{target\_region}: A list specifying the target(s) of the inquiry. Each item in this list defines which image is used and where on that image the inquiry is applied.}
        \begin{itemize}
            \item \texttt{\textbf{image\_index}: Identifies the target image.}
            \item \texttt{\textbf{bbox\_2d}: A list of four integer coordinates `[x1, y1, x2, y2]`...}
        \end{itemize}
        \item \texttt{\textbf{Inquiry}: This is the semantic goal of the code/box. It must describe the information the code/box is trying to highlight, verify, or expose within the specified `bbox\_2d`. It is the `why' behind the crop, not just the act of cropping.}
    \end{itemize}

    \textbf{Output Schema:}
    \begin{verbatim}
[
    {
        "thought":  "The reasoning process leading to the target regions and inquiry",
        "target_region": [{
            "image_index": "The index of the first image...",
            "bbox_2d": [x1, y1, x2, y2]
        }, ...],
        "inquiry": "The inferred purpose of the corresponding code block or bounding box.",
    }, ...
]
    \end{verbatim}
    \end{tcolorbox}

    \caption{
        The detailed prompt structure used to translate heterogeneous, tool- or code-based reasoning trajectories into a unified, structured JSON format for SFT. }
    \label{fig:full_translation_prompt}
\end{figure*}

\subsection{Component 2: Synthesizing Spatial Reasoning Trajectories}
\label{sec:synthesis_pipeline}
The second SFT data component targets complex spatial reasoning datasets~\citep{zhang2025flatland,ouyang2025spacerreinforcingmllmsvideo,feng2025visuospatial}. These datasets require models to deduce intricate spatial and temporal relationships but lack annotated reasoning trajectories. A common practice to address this is rejection sampling~\cite{touvron2023llama2}, where a teacher LVLM generates entire trajectories that are then filtered based on the final answer. We argue this method is sub-optimal for our needs, as a static teacher model cannot produce the rich, text-conditioned, multi-region manipulations that our framework is designed for.

Therefore, we designed a dedicated synthesis pipeline, illustrated in Figure~\ref{fig:synthesis_pipeline}, to generate high-quality ``chatting with images'' trajectories. This pipeline first programmatically mines latent spatial and contextual knowledge before synthesizing the final reasoning path. The steps are as follows:
\begin{figure}[t]
    \centering
    \includegraphics[width=\columnwidth]{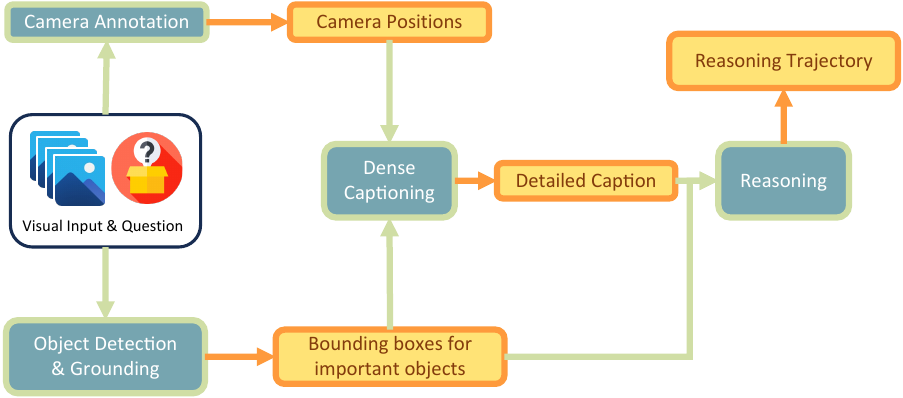} 
    \caption{The dedicated pipeline for synthesizing high-quality spatial reasoning trajectories. We first mine latent knowledge (objects, grounding, camera motion, captions) and then use this rich context to guide a powerful teacher model (Qwen2.5-VL-72B) in generating a detailed reasoning path.}
    \label{fig:synthesis_pipeline}
\end{figure}

\begin{enumerate}
    \item \textbf{Input}: The pipeline starts with the initial visual input (images/video) and the user's question.
    \item \textbf{Parallel Knowledge Mining}: Several analyses are performed in parallel to extract multi-faceted information:
    \begin{itemize}
        \item \textbf{Object detection and grounding}: Based on the question, Qwen2.5-VL-72B identifies all potentially important objects in the scene. For each identified object, Qwen2.5-VL-72B generates its corresponding bounding box (``bbox'') for each frame it appears in.
        \item \textbf{Camera annotation}: We use VGGT~\citep{wang2025vggt} to estimate camera motion across input images, providing crucial information about viewpoint changes. This yields a sequence of camera positions.
    \end{itemize}
    \item \textbf{Holistic Captioning}: Qwen2.5-VL-72B generates a single, comprehensive caption for the entire visual input. This caption describes not only the static objects and their relationships but also the temporal dynamics, such as camera movements between different frames.
    \item \textbf{Reasoning Generation}: All the mined information, including important objects and their bounding boxes, camera positions, and detailed captions, is consolidated and fed as rich context to Qwen2.5-VL-72B. With this comprehensive understanding of the scene, the model is prompted to generate a high-quality reasoning trajectory (interleaving thoughts and tool calls) and the final answer.
    \item \textbf{Output}: The final output from this guided process is a complete ``chatting with images'' trajectory generated by Qwen2.5-VL-72B. This trajectory, which includes the detailed reasoning path and the final outcome, serves as a gold-standard training sample for our model.
\end{enumerate}

This structured, knowledge-first approach ensures the generated trajectories are not only correct but also logically sound and grounded in detailed visual evidence, teaching our model to perform nuanced visual reasoning.

% You should replace 'your_image_file' with the actual filename of your diagram.

\subsection{Component 3: Augmenting with Text-Only Trajectories}
Finally, to ensure our model can efficiently handle simpler questions where extensive visual re-examination is unnecessary, we augment the SFT dataset with purely textual reasoning trajectories. These ``thinking about images'' trajectories teach the model to answer questions when the visual information from the initial encoding is self-evident~\citep{yang2025kwai}. This component is sourced in two ways:
\begin{itemize}
    \item For datasets that already provide text-only reasoning chains~\cite{zhang2025thyme,wang2025traceableevidenceenhancedvisual}, we use them directly.
    \item For datasets without such annotations, we use Qwen2.5-VL-72B to generate reasoning chains via rejection sampling~\cite{dubey2024llama}. We generate multiple candidate trajectories for each question and retain only those that lead to the correct final answer, ensuring the quality of the synthesized data.
\end{itemize}
This hybrid data strategy equips \textsc{ViLaVT} with the ability to dynamically choose between deep, iterative visual thinking and efficient, text-based reasoning, depending on the complexity of the query.

%operationalize this by
%This allows us to convert a wide array of such examples into our native format of ``chatting with images.'' Specifically, we 

\section{Benchmark Details}\label{benchmark_details}

\begin{table*}[!t]
    \centering
    \caption{Statistics of examples in high-resolution and spatial reasoning benchmarks.}
    \resizebox{\linewidth}{!}{
    \begin{tabular}{@{}cccccccccc@{}}
        \toprule
        \multicolumn{2}{c}{\textbf{High-Resolution}} & \multicolumn{6}{c}{\textbf{Spatial Reasoning}} \\
        \cmidrule(lr){1-2} \cmidrule(lr){3-8}
        
        \multicolumn{2}{c}{\textbf{Single-Image}} & \multicolumn{2}{c}{\textbf{Single-Image}} & \multicolumn{3}{c}{\textbf{Multi-view}} & \multicolumn{1}{c}{\textbf{Video}} \\
        \cmidrule(lr){1-2} \cmidrule(lr){3-4} \cmidrule(lr){5-7} \cmidrule(lr){8-8}
        
        \textbf{HRBench-4K} & \textbf{HRBench-8K} & 
        \textbf{SpatialEval-Real} & \textbf{EmbSpatial} &
        \textbf{ERQA} & \textbf{SPAR-Bench} & \textbf{MMSI-Bench} &
        % \textbf{ViewSpatial} &
        \textbf{VSI-Bench} \\
        \midrule
        
        800 & 800 &
        135 & 3,640 & 
        400 & 7,211 & 1,000 &
        % 5,712 &
        5,130 \\
        
        \bottomrule
    \end{tabular}}
    \label{tab:benchmark}
\end{table*}

\begin{table*}[t]
  \centering
  \caption{Comparative per-round overhead analysis of different reasoning paradigms. For interactive methods, we report the average latency of a single reasoning--interaction round, since different paradigms may require varying numbers of rounds. 
  Our paradigm achieves high flexibility with a controlled overhead.
  % avoiding the severe latency of external execution loops.
  }
  \label{tab:overhead}
  \resizebox{\linewidth}{!}{%
  \begin{tabular}{@{}llcc@{}}
    \toprule
    \textbf{Paradigm} & \textbf{Model Example} & \textbf{Relative Latency} & \textbf{Key Overhead Source} \\ %& \textbf{End-to-End Differentiable} 
    \midrule
    Static (Non Thinking) & Qwen2.5-VL & 2.51s (-) & Single forward pass  \\
    Static (Thinking \textit{about} Images) & Qwen2.5-VL & 4.69s (1.0$\times$) & Single forward pass  \\
    Tool-based (Thinking \textit{with} Images) & ViLaSR~\citep{wu2025reinforcing} & 6.05s (1.28$\times$) & External tool calls, I/O  \\
    Code-based (Thinking \textit{with} Images) & Thyme~\citep{zhang2025thyme} & 6.18s (1.31$\times$) & Code interpreter, process isolation, I/O  \\
    \textbf{Ours} (Chatting \textit{with} Images) & VILAVT (Ours) & \textbf{5.97s (1.06$\times$)} & Internal feature re-computation \\

    \bottomrule
  \end{tabular}%
  }
\end{table*}

\begin{table*}[b]
    \centering
    \caption{Detailed Results on VSI-Bench.}
    \label{tab:vsibench}
    \resizebox{\linewidth}{!}{
    \begin{tabular}{l ccccm{0.01em}cccc c}
        \toprule
        \multirow{4}{*}{\textbf{Method}} & \multicolumn{9}{c}{\textbf{Sub-task}s} & \multirow{4}{*}{\textbf{Average}} \\
        \cmidrule{2-10}
        & \multicolumn{4}{c}{\textbf{Numerical questions}} & & \multicolumn{4}{c}{\textbf{Multiple-choice questions}} \\
        \cmidrule{2-5}
        \cmidrule{7-10}
        & Object & Absolute & Object & Room & & Relative & Relative & Route & Appearance \\
        & Count & Distance & Size & Size & & Distance & Direction & Plan & Order \\
        \midrule
        \midrule
        \multicolumn{10}{c}{\textbf{Proprietary LVLMs}} \\
        \midrule
        GPT-4o & 46.2 & 5.3 & 43.8 & 38.2 & & 37.0 & 41.3 & 31.5 & 28.5 & 34.0 \\
        Gemini-1.5-Flash & 49.8 & 30.8 & 53.5 & 54.4 & & 37.7 & 41.0 & 31.5 & 37.8 & 42.1 \\
        Gemini-1.5-Pro & 56.2 & 30.9 & 64.1 & 43.6 & & 51.3 & 46.3 & 36.0 & 34.6 & 45.4 \\
        \midrule
        \midrule
        \multicolumn{10}{c}{\textbf{Open-source LVLMs}} \\
        \midrule
        Qwen2.5-VL-7B & 29.3 & 26.8 & 54.6 & 35.5 & & 41.0 & 34.4 & 26.8 & 29.3 & 34.7 \\
        Qwen2.5-VL-72B & 33.9 & 27.2 & 59.3 & 28.5 & & 47.2 & 35.3 & 22.2 & 34.5 & 36.0 \\
        LLaVA-NeXT-Video-7B & 48.5 & 14.0 & 47.8 & 24.2 & & 43.5 & 42.4 & 34.0 & 30.6 & 35.6 \\
        LLaVA-OneVision-7B & 47.7 & 20.2 & 47.4 & 12.3 & & 42.5 & 35.2 & 29.4 & 24.4 & 32.4 \\
        Kimi-VL-A3B-Instruct-16B & - & - & - & - & & - & - & - & - & 37.4 \\
        LLaVA-NeXT-Video-72B & 48.9 & 22.8 & 57.4 & 35.3 & & 42.4 & 36.7 & 35.0 & 48.6 & 40.9 \\
        LLaVA-OneVision-72B & 43.5 & 23.9 & 57.6 & 37.5 & & 42.5 & 39.9 & 32.5 & 44.6 & 40.2 \\
        InterVL2-8B & 31.3 & 29.0 & 48.9 & 44.2 & & 38.0 & 33.4 & 28.9 & 46.4 & 37.5 \\
        \midrule
        \midrule
        \multicolumn{10}{c}{\textbf{Representative methods for spatial reasoning}} \\
        \midrule
        SpaceR-7B & 62.3 & 31.8 & 60.5 & 40.2 & & 42.1 & 46.8 & 32.0 & 48.2 & 45.5 \\
        Spatial-MLLM-4B & 65.3 & 34.8 & 63.1 & 45.1 & & 41.3 & 46.2 & 33.5 & 46.3 & 48.4 \\
        ViLaSR-7B & 63.5 & 34.4 & 60.6 & 30.9 & & 48.9 & 45.2 & 30.4 & 49.2 & 45.4 \\
        \midrule
        \midrule
        \multicolumn{10}{c}{\textbf{Ours}} \\
        \midrule
        \textsc{ViLavT}-SFT & 35.9 & 32.8 & 61.4 & 39.8 & & 43.7 & 42.2 & 29.9 & 60.7 & 43.3 \\
        % \textsc{ViLaVT} vanilla ViT & 63.9  & 32.9	& 64.3	& 41.1 & & 42.4	& 45.4	& 29.9	& 59.4 & 47.4 \\
        % \textsc{ViLavT} & 64.0	&40.2	&70.2	& 50.3 & & 46.6 & 44.5	& 28.9	& 65.7 & 51.3 \\
        \textsc{ViLavT} & 65.8	& 39.0	& 69.2	& 55.5 & & 50.8 & 46.0	& 23.7	& 66.3 & 52.0 \\
        \bottomrule
    \end{tabular}}
\end{table*}

We conduct comprehensive evaluations across two categories of benchmarks, each targeting distinct capabilities of vision-language models.

\textbf{High-Resolution Perception Benchmarks.} These benchmarks assess a model's ability to compress high-resolution visual information while maintaining fine-grained detail perception and conducting precise visual search—tasks that pose significant challenges to current VLMs. We employ \textit{HRBench-4K} and \textit{HRBench-8K}~\citep{wang2025divide}, which evaluate model performance on 4K and 8K resolution images, respectively.

\textbf{Spatial Reasoning Benchmarks.} These benchmarks evaluate the model's capacity to understand complex spatial and temporal relationships encoded in visual data. We organize them by input modality: 
\begin{itemize}[leftmargin=*,nosep]
    \item \textbf{Single-image reasoning:} \textit{SpatialEval-Real}~\citep{wang2024isapicture} and \textit{EmbSpatial}~\citep{du-etal-2024-embspatial} test spatial understanding within individual images.
    \item \textbf{Multi-image reasoning:} \textit{ERQA}~\citep{team2025gemini}, \textit{SPAR-Bench}~\citep{zhang2025flatland}, and \textit{MMSI-Bench}~\citep{yang2025mmsibenchbenchmarkmultiimagespatial} evaluate cross-image spatial reasoning.
    \item \textbf{Video-based reasoning:} \textit{VSI-Bench}~\citep{yang2024think} 
    % and \textit{ViewSpatial}~\citep{li2025viewspatialbenchevaluatingmultiperspectivespatial} 
    assess spatiotemporal understanding in video sequences.
\end{itemize}

Table~\ref{tab:benchmark} summarizes the dataset statistics across all benchmarks.

\section{Implementation Details}\label{implementation_details}

\begin{itemize}
    \item \textbf{Vision encoder architecture.} 
    The re-invented vision Transformer $\mathcal{V}_e$ in \textsc{ViLaVT} comprises $32$ Transformer layers with $16$ attention heads and a hidden dimension of $1280$, inherited from Qwen2.5-VL~\citep{bai2025qwen2}. The lightweight language Transformer $\mathcal{V}_m$ is a frozen Qwen3-Embedding-0.6B model.\footnote{\url{https://huggingface.co/Qwen/Qwen3-Embedding-0.6B}} 
    We employ a hybrid attention mechanism across different layers:
    \begin{itemize}[leftmargin=*, itemsep=1pt, topsep=2pt]
        \item \textit{Intra-full attention} (layers 8, 16): Attention is operated within individual images.
        \item \textit{Inter-full attention} (layers 17--32): Full self-attention jointly operates over all visual tokens from multiple images and the textual inquiry, enabling query-conditioned cross-view interactions.
        \item \textit{Sparse window attention} (remaining layers): attention is confined to local patch windows for efficiency.
        \item \textit{Sparse window attention} (remaining layers): Attention is confined to local patch windows for computational efficiency.
    \end{itemize}

    \item \textbf{SFT.}
    In the SFT stage, we train for 2 epochs with a global batch size of 384, using AdamW~\citep{loshchilov2018decoupled} with a learning rate of $5\times10^{-5}$.

    \item \textbf{RL.}
    The RL stage follows the VeRL framework~\citep{sheng2024hybridflow}. For each instance, we generate $G=4$ candidate trajectories per rollout.
    A rollout terminates when a final answer is produced, when the number of interaction rounds exceeds $T_{\max}$ (5 for general tasks and 10 for spatial reasoning), or when a cumulative limit of 52 processed visual inputs is reached. The policy is updated with a batch size of 96, and $\epsilon_1/\epsilon_2/\delta$ in Eq.~\ref{rl_object} are set to 0.2/0.3/$10^{-6}$, respectively.
    
    \item \textbf{Hardware and training infrastructure.} 
    All experiments are conducted on a cluster of $96$ NVIDIA H20 (80GB) GPUs. The SFT stage requires approximately $27$ hours, while the RL stage takes roughly $138$ hours under this configuration. 
    
    % \item \textbf{Data preprocessing.} 
    % Visual inputs are processed with resolution-aware strategies:
    % \begin{itemize}[leftmargin=*, itemsep=1pt, topsep=2pt]
    %     \item \textit{High-resolution images}: Resized to a maximum resolution of $8192 \times 28 \times 28$.
    %     \item \textit{Videos}: We uniformly sample $32$ frames per clip, with each frame processed at up to $256 \times 28 \times 28$.
    %     \item \textit{Multi-view inputs}: Total visual tokens are constrained to a fixed budget of $8{,}192$ tokens, dynamically allocating resolution based on the number of views.
    % \end{itemize}
    
    \item \textbf{Inference optimization.} 
    During evaluation and RL rollouts, we leverage vLLM~\citep{kwon2023efficient} for batched inference with optimized memory management. We use top-$p$ sampling~\citep{holtzmancurious} ($p=0.9$) with temperature $0.75$ to generate reasoning trajectories and final answers.
    We adapt \textsc{ViLaVT} to the vLLM framework and will open-source our implementation upon paper acceptance.
    
    \item \textbf{Evaluation protocol.} 
    We impose task-specific limits on tool usage: up to $10$ tool calls for spatial reasoning tasks and $5$ calls for high-resolution tasks. All high-resolution benchmarks are evaluated using VLMEvalKit~\citep{duan2024vlmevalkit} to ensure consistency with baseline methods. Final answers are extracted from \texttt{<answer>} tags and normalized before comparison.
\end{itemize}

\section{More Analysis}\label{ablation_study}

\subsection{Analysis of Training Dynamics}
\label{sec:training_dynamics}

\begin{figure}[htbp]
    \centering
    \begin{subfigure}[b]{0.48\linewidth}
        \centering
        \includegraphics[width=\linewidth]{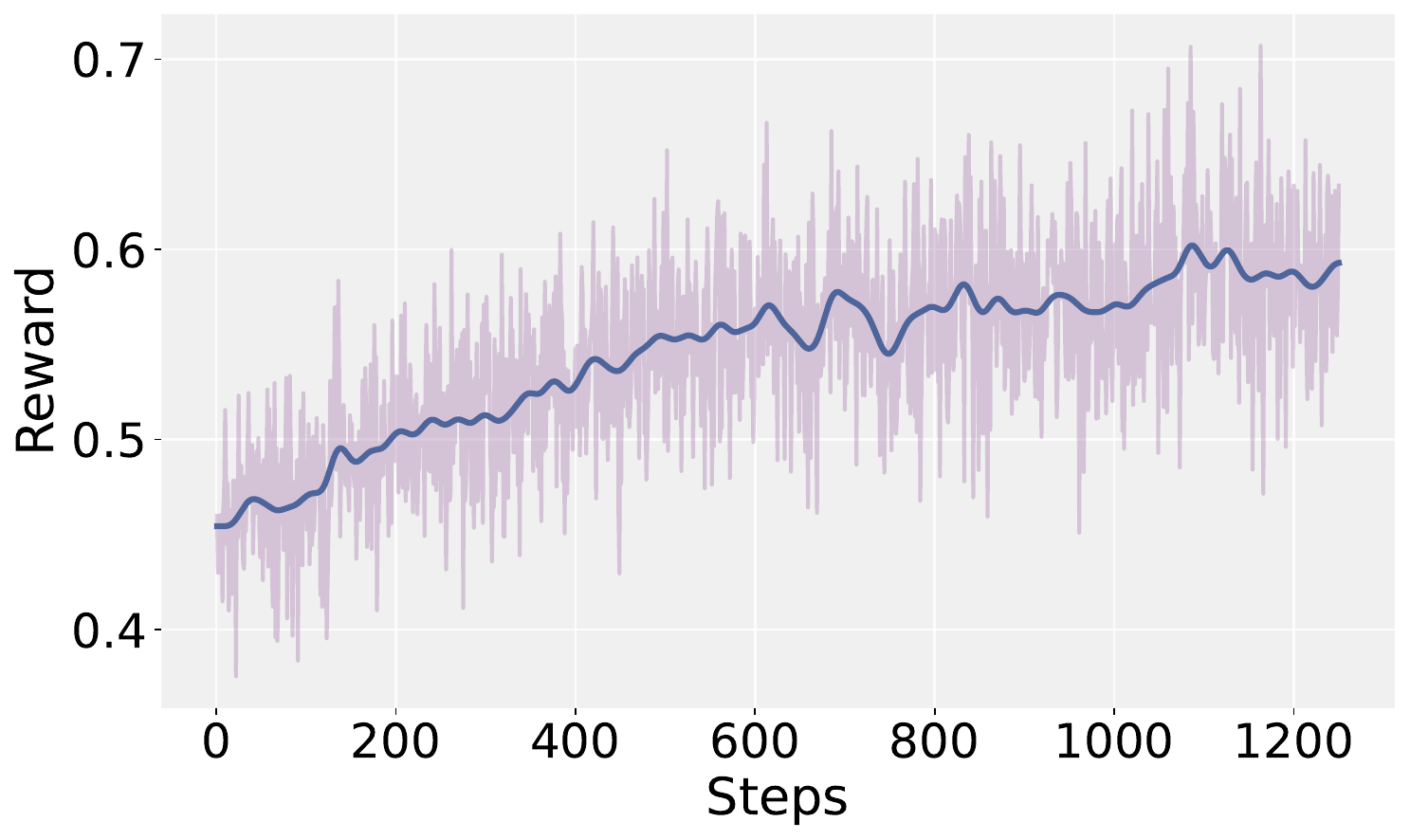}
        \caption{Overall reward $R$}
        \label{fig:reward}
    \end{subfigure}
    \hfill
    \begin{subfigure}[b]{0.48\linewidth}
        \centering
        \includegraphics[width=\linewidth]{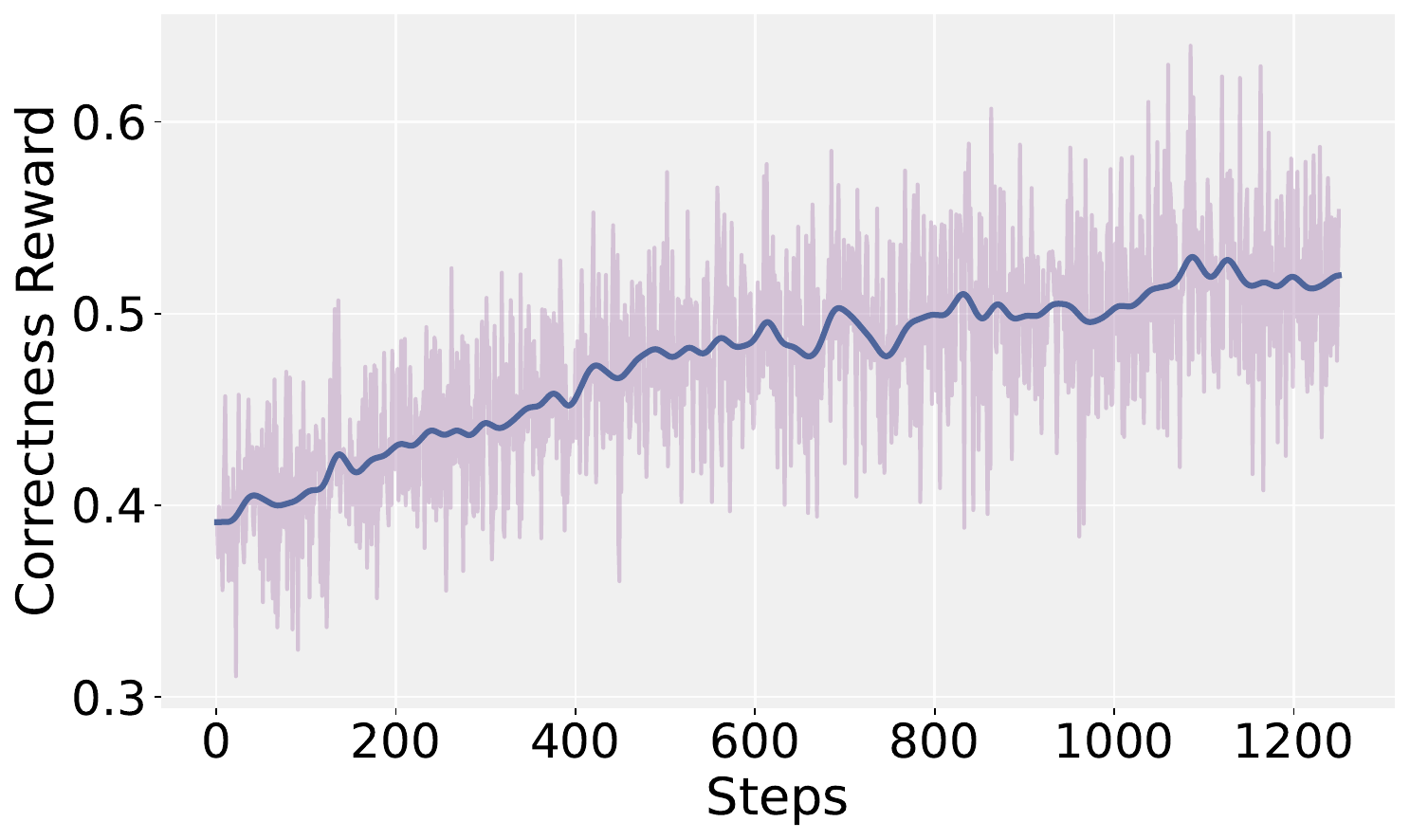}
        \caption{ Correctness reward $R_\text{correct}$}
        \label{fig:acc_reward}
    \end{subfigure}
    
    \vspace{1em}
    
    \begin{subfigure}[b]{0.48\linewidth}
        \centering
        \includegraphics[width=\linewidth]{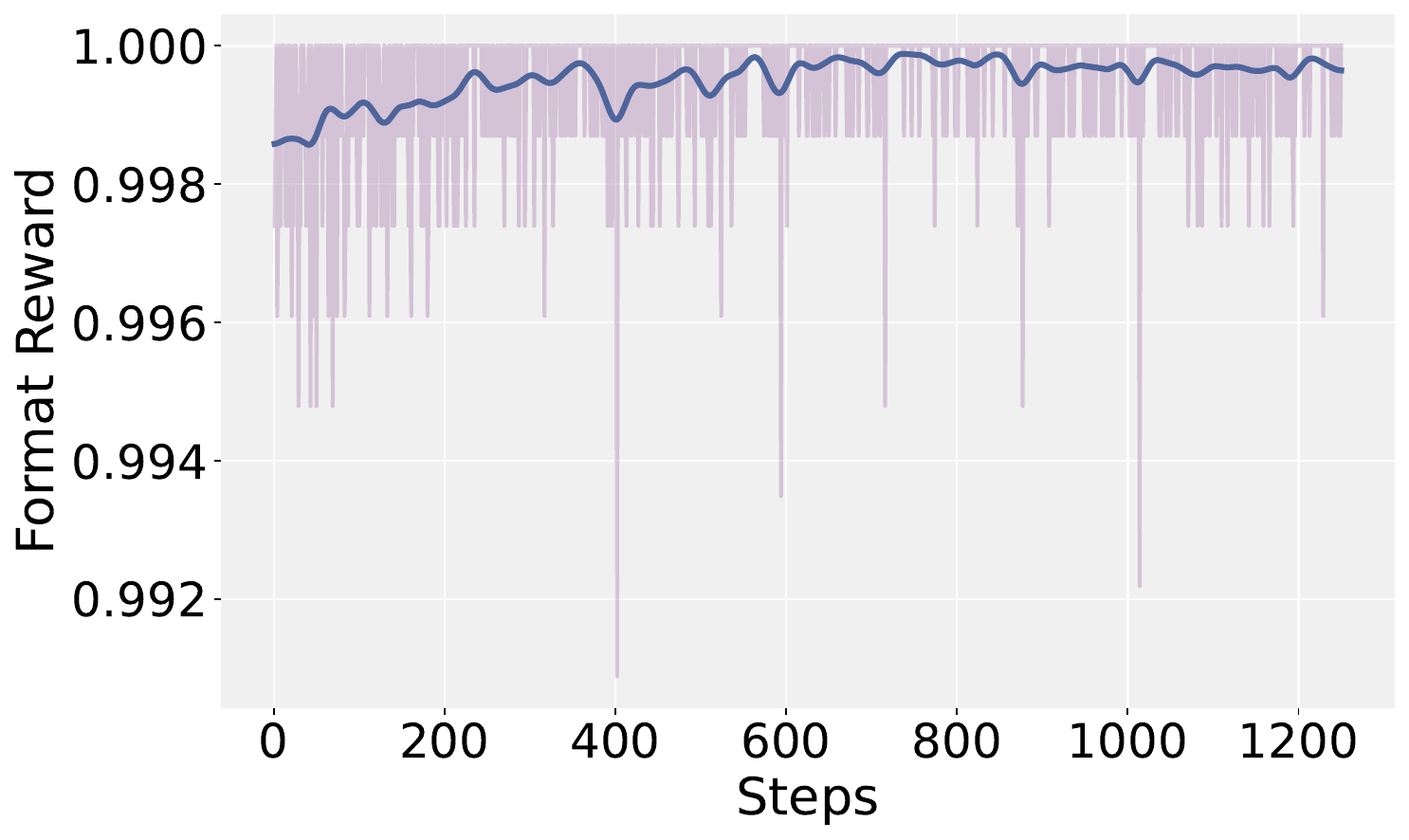}
        \caption{Format reward $R_\text{format}$}
        \label{fig:format_reward}
    \end{subfigure}
    \hfill
    \begin{subfigure}[b]{0.48\linewidth}
        \centering
        \includegraphics[width=\linewidth]{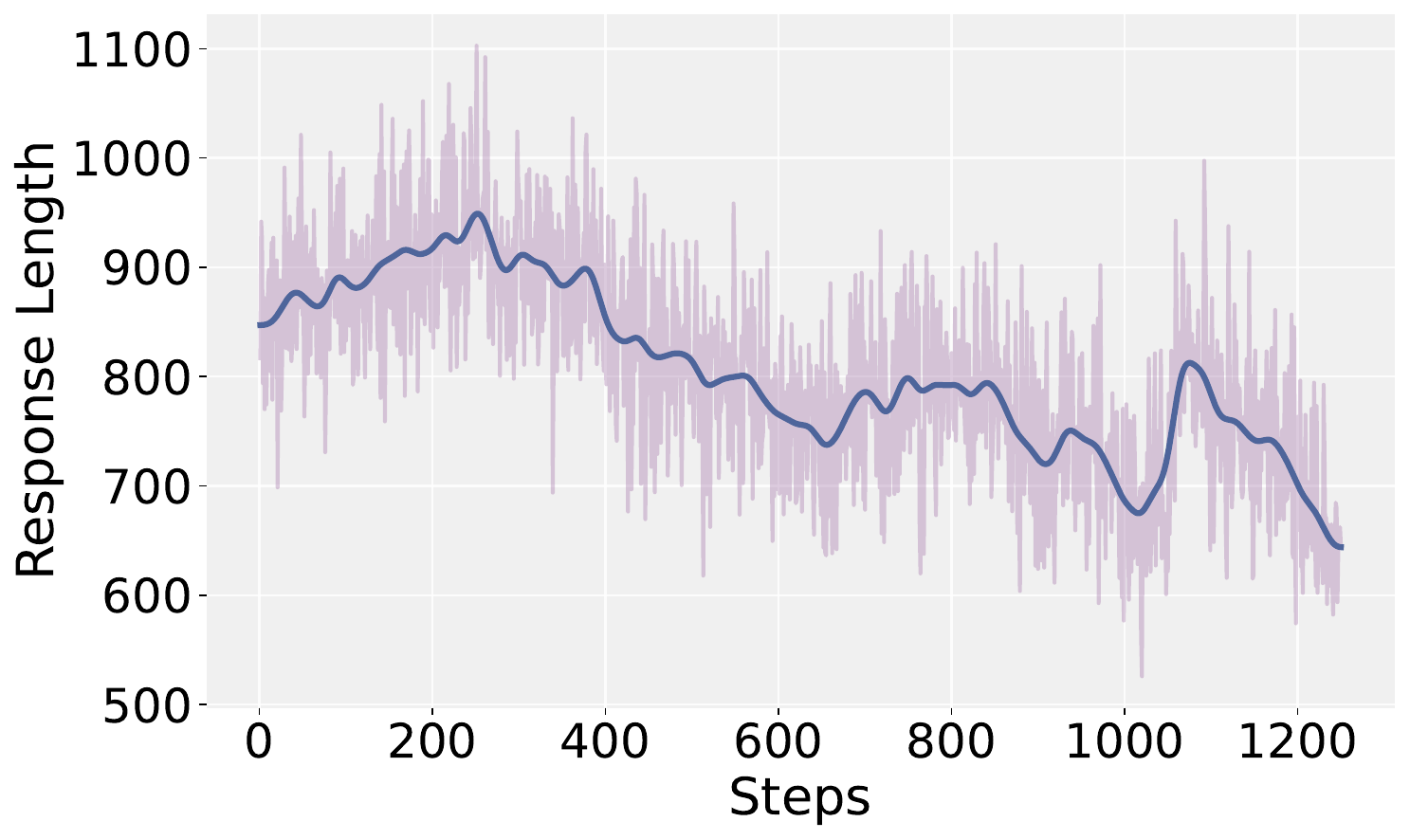}
        \caption{Response length $N$}
        \label{fig:response_length}
    \end{subfigure}
    
    \caption{RL training dynamics of \textsc{ViLaVT}}
    \label{fig:training_dynamics}
\end{figure}

Figure~\ref{fig:training_dynamics} illustrates the training dynamics of \textsc{ViLaVT}. We observe that both the correctness reward $R_\text{correct}$ and the overall reward $R$ exhibit a stable upward trend, indicating continuous optimization throughout training. The format reward $R_\text{format}$ starts relatively high due to extensive supervised fine-tuning, which endows the model with strong format-following capabilities;  it steadily approaches 1.0 as training progresses. It is noteworthy that the response length $N$ increases over the first 200 training steps, peaks, and then gradually declines. We hypothesize that this reflects the emergence of novel reasoning patterns in \textsc{ViLaVT} that differ from those present in the supervised data. In later RL iterations, $N$ shows a decreasing trend but with noticeable fluctuations.

\subsection{Overhead Analysis}\label{overhead_analysis}

Our “chatting with images’’ paradigm intentionally trades a controlled computational overhead for substantial gains in reasoning accuracy. We analyze this trade-off in Table~\ref{tab:overhead}, comparing our model's per-round inference-stage overhead against both ``thinking about image'' and ``thinking with image'' paradigms on \textit{HRBench-4K}. Since different interactive paradigms may require varying numbers of reasoning rounds, total inference time is not directly comparable; instead, we report the average cost of a single reasoning–interaction step. All timing measurements are recorded using standard Transformer inference without acceleration techniques.

As the analysis shows, while interactive reasoning is inherently more costly than a single-pass baseline, our per-round cost is comparable to, and slightly lower than, tool-based and code-based approaches. This efficiency stems from using a simple image crop operation and a unified manipulation interface via inquiry-conditioned, multi-image re-encoding, avoiding external tool I/O and code execution loops. In addition, (i) the inquiry encoder $\mathcal{V}_m$ is a small frozen model with negligible overhead, and (ii) re-computation is applied only to cropped regions rather than the full-resolution input. Together, these choices yield a modest per-step overhead while remaining competitive with prior interactive paradigms.

% As the analysis shows, while any interactive process is inherently more costly than a single-pass static model, our approach is significantly more efficient than code-based methods. This is because our entire reasoning loop is a unified, end-to-end forward pass within a single model, avoiding the high-latency bottlenecks of external process calls, code execution, and data I/O that plague programmatic approaches. This efficiency is achieved by design: (1) the inquiry encoder ($V_m$) is a small, frozen language model, adding negligible cost; and (2) re-computation is performed only on small, cropped image regions, not the full-resolution input. In essence, our framework makes a principled trade-off, accepting a modest, per-step overhead to unlock a far more powerful and flexible reasoning capability that remains more efficient and scalable than prior interactive paradigms.

\subsection{Detailed results on VSI-Bench}
In this section, we present detailed results of \textsc{ViLaVT} on \textit{VSI-Bench}~\citep{yang2024think}, a challenging benchmark covering eight diverse spatial reasoning tasks. Table~\ref{tab:vsibench} compares our method with state-of-the-art models across both numerical questions (object count, absolute distance, object size, room size) and multiple-choice questions (relative distance, relative direction, route planning, appearance order). The results highlight the comprehensive spatial reasoning capabilities of our approach.

\section{Case Study}\label{app:more_cases}

\subsection{Representative Cases}

\begin{figure*}[!t]
    \centering
    \includegraphics[width=0.96\textwidth]{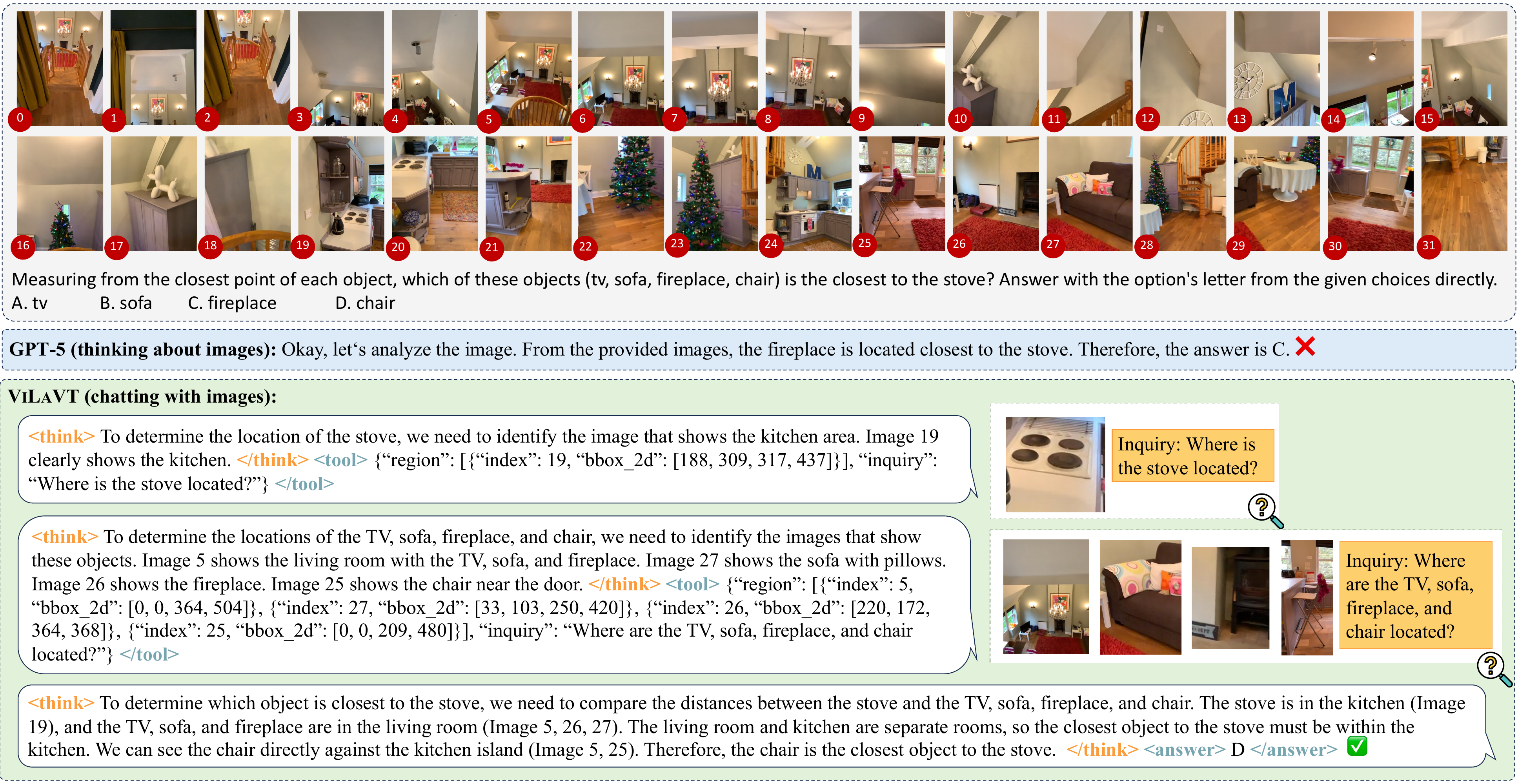}
    \caption{A comparison on a multi-frame spatial reasoning task from VSI-Bench.}
    \label{fig:case_study}
\end{figure*}

Figure~\ref{fig:case_study} demonstrates our paradigm's strength on video-based spatial reasoning. 

\begin{figure*}
    \centering
    \includegraphics[width=0.8\textwidth]{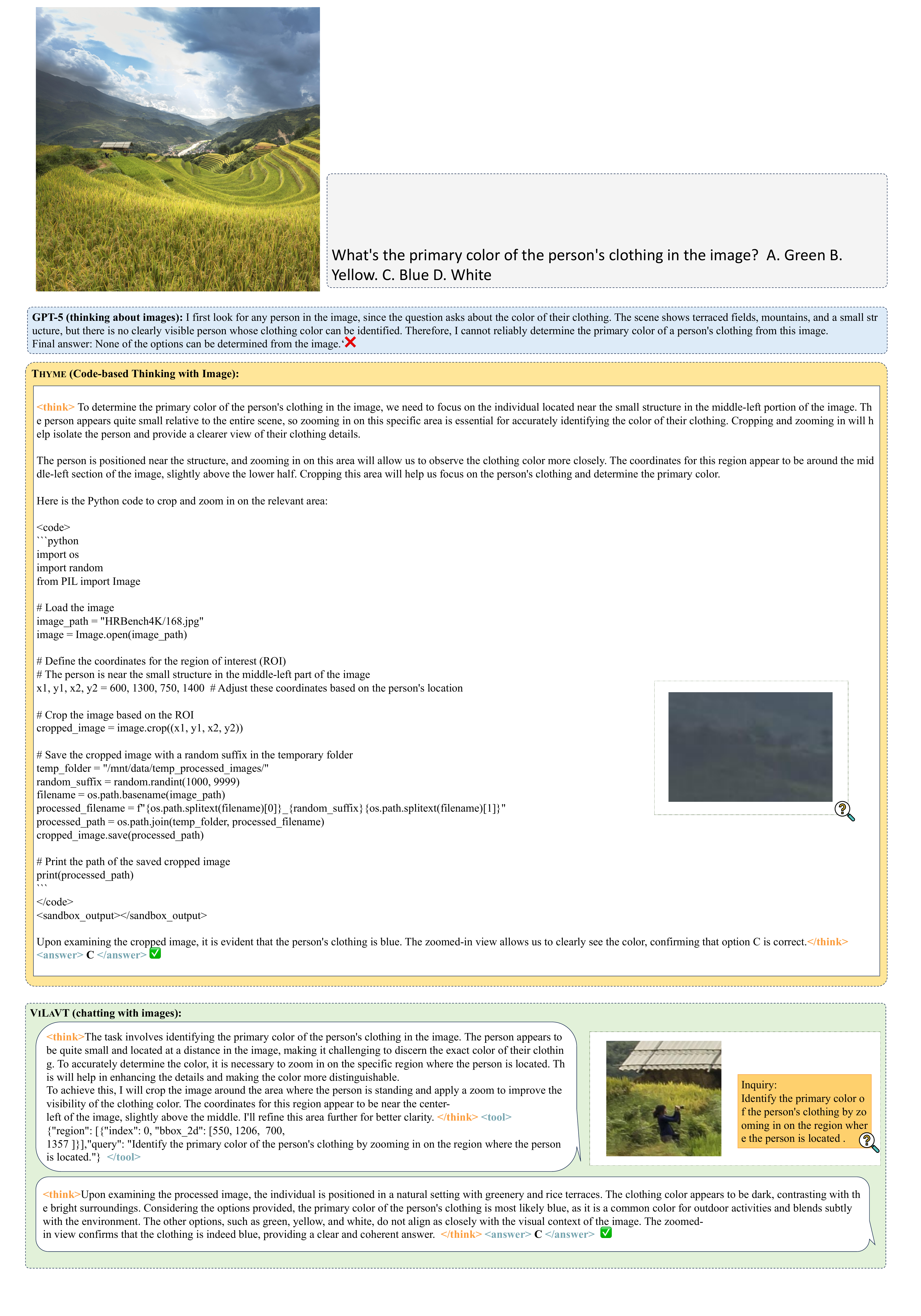}
    \caption{First case (\textit{HRBench-4K}, high-resolution perception). \textsc{ViLaVT} first analyzes the entire image and then crops and zooms in on the small regions containing the person, ultimately correctly concluding that the person’s clothing is blue. Although Thyme invokes code execution, it fails to crop the correct region.
}
    \label{fig:case_1}
\end{figure*}

\begin{figure*}
    \centering
    \includegraphics[width=\textwidth]{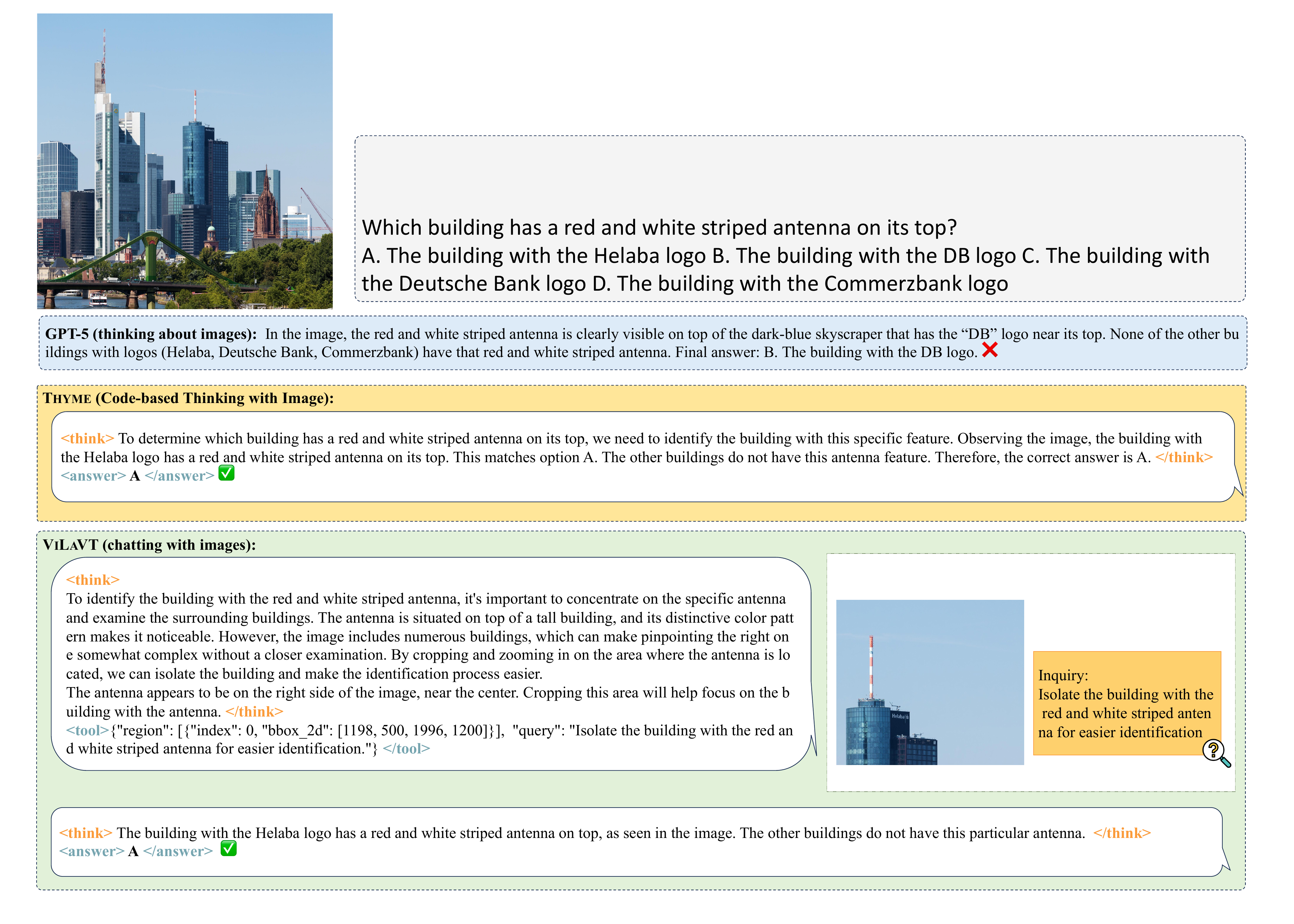}
    \caption{Second case (\textit{HRBench-4K}, high-resolution perception). \textsc{ViLaVT} first performs targeted cropping and then correctly infers that the building has a red and white striped antenna. GPT-5 directly produces an incorrect answer, while Thyme outputs a response but does not provide an explicit reasoning process.}
    \label{fig:case_2}
\end{figure*}

\begin{figure*}
    \centering
    \includegraphics[width=\textwidth]{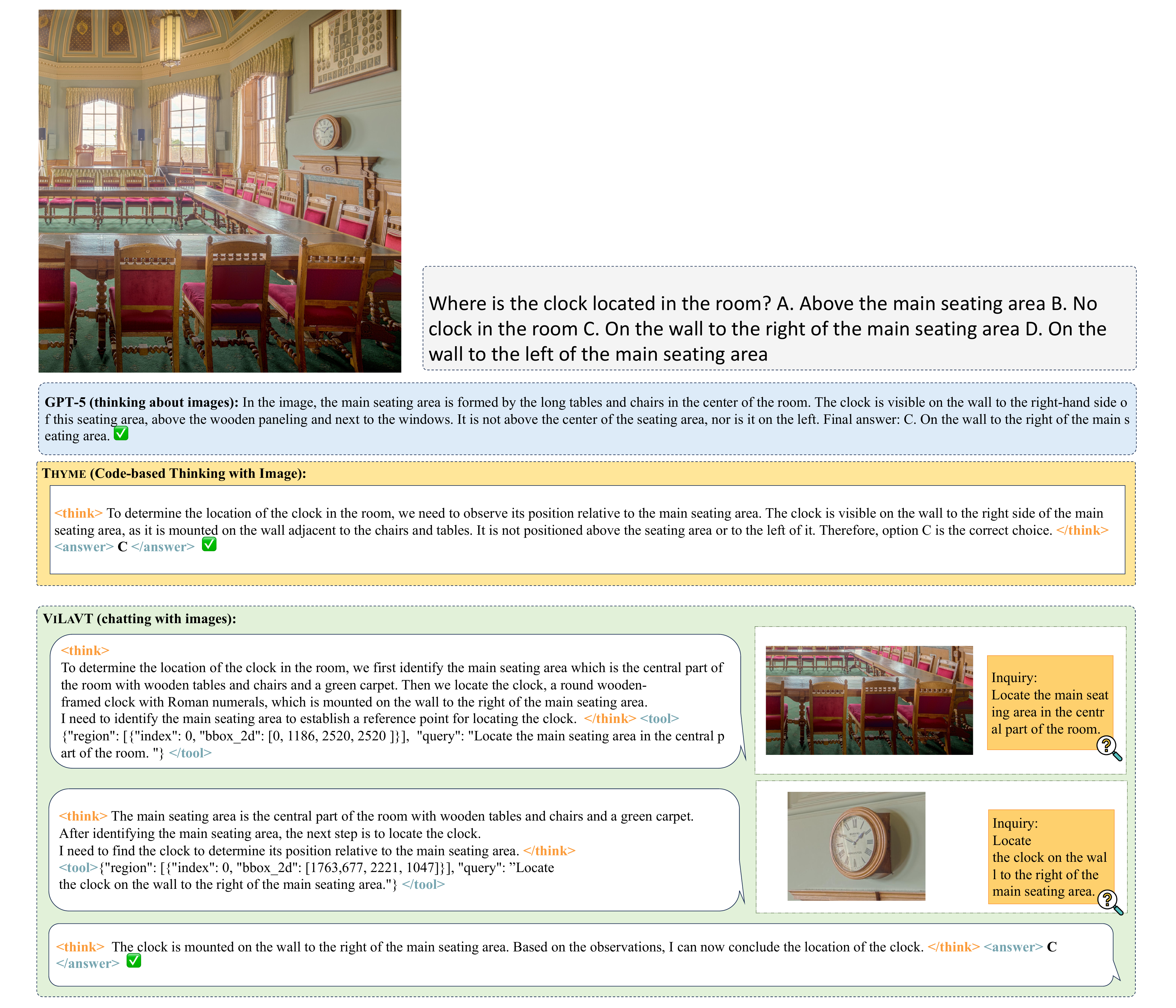}
    \caption{Third case (\textit{HRBench-4K}, high-resolution perception). For reasoning about relative object positions, \textsc{ViLaVT} explicitly grounds the entities in the image and then derives the final answer. Both GPT-5 and Thyme arrive at the correct answer directly through text-based reasoning.
    } 
    \label{fig:case_3}
\end{figure*}

There are also three HRBench-4K cases that together highlight the strengths and weaknesses of different models in high-resolution visual perception and reasoning. In Figure~\ref{fig:case_1}, \textsc{ViLaVT} leverages a coarse-to-fine strategy—first scanning the full image, then cropping and zooming into small regions—to correctly identify that the person’s clothing is blue, whereas Thyme gives an incorrect cropping region. In Figure~\ref{fig:case_2}, \textsc{ViLaVT} again benefits from targeted cropping to correctly recognize the building’s red and white striped antenna, while GPT-5 directly outputs an incorrect answer and Thyme provides an answer without an explicit reasoning chain. In contrast, the example in Figure~\ref{fig:case_3} shows that when the task mainly involves reasoning about relative spatial relationships, all three models can arrive at the correct answer, with \textsc{ViLaVT} explicitly grounding entities in the image and the others relying primarily on text-based reasoning.

\subsection{Attention Map Visualization Cases}
\label{app:visualization_cases}

We extract attention maps from the last layer of the vision encoder, average them across all heads, and aggregate patch-level attention to form a single heatmap, which is overlaid on the original image for visualization.

\begin{figure*}
    \centering
    \includegraphics[width=\textwidth]{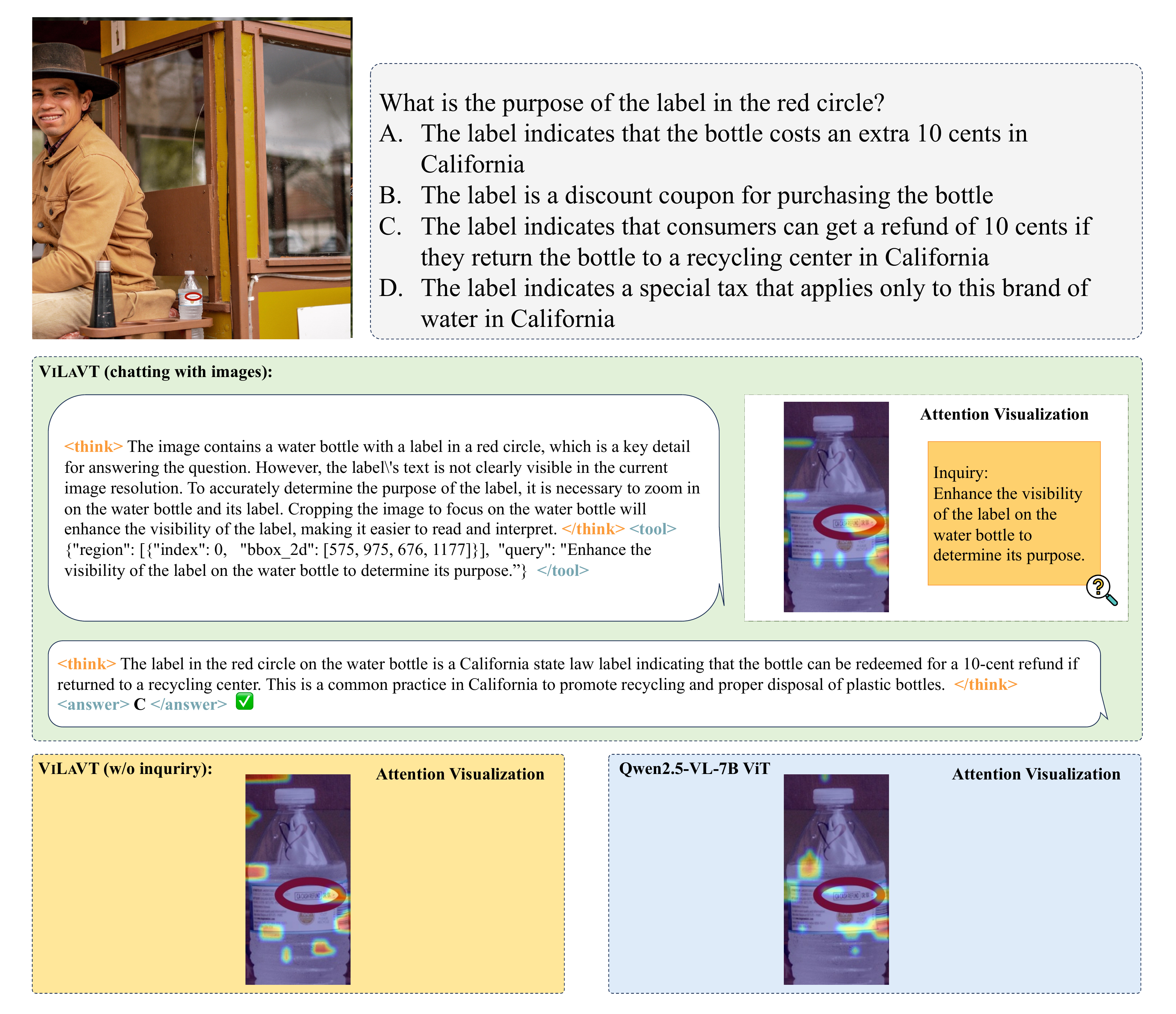}
    \caption{Attention map visualization on an example from HRBench-4K.}
    \label{fig:attn_case_578}
\end{figure*}

We provide another attention map visualization case in Figure~\ref{fig:attn_case_578}. Compared with \textsc{ViLaVT} w/o Inquiry and Qwen2.5-VL-7B with the vanilla ViT backbone, the query-aware vision encoder of \textsc{ViLaVT} focuses more strongly on the query-relevant object—the label in the red circle. This indicates that incorporating the textual query into the vision encoder facilitates more fine-grained detail extraction.

% In the spatial reasoning case shown in Figure~\ref{fig:attn_spar_case3361}, we find that in the first reasoning step \textsc{ViLaVT} selects a frame region containing the “red point” as the reference object. In the next step, it jointly attends to both the “red point” and the “blue point” to infer the depth of the blue point, and the corresponding attention map adjusts accordingly. Compared with \textbf{\textsc{ViLaVT} w/o Inquiry}, \textsc{ViLaVT} exhibits more pronounced and interpretable attention shifts in response to changes in the query and the input images. In contrast, the vanilla ViT in Qwen2.5-VL-7B produces identical attention maps for the “red point” in both single-image and multi-image settings, indicating weaker adaptation to the query and visual context.

% \begin{figure*}
%     \centering
%     \includegraphics[width=\textwidth]{image/attn_spar_case3361.pdf}
%     \caption{Attention map visualization on a spatial reasoning example.    } 
%     \label{fig:attn_spar_case3361}
% \end{figure*}

\section{Prompt Template}
\label{app:prompt}
The prompting framework of \textsc{ViLaVT} comprises a system prompt that encodes reasoning principles (Figure~\ref{fig:sys_prompt}) and a user prompt template that structures visual inputs and tool interactions (Figure~\ref{fig:user_prompt}).

\begin{figure*}[t!]
    \centering
    % System Prompt Box
    % \begin{tcolorbox}[
    %     colback=blue!5,
    %     colframe=blue!75!black,
    %     width=\textwidth,
    %     arc=2mm,
    %     auto outer arc,
    %     title=\textbf{System Prompt: Guiding Principles for ``Thinking with Images'' Data Conversion},
    %     fonttitle=\bfseries,
    %     boxsep=2mm,
    %     toptitle=2mm,
    %     bottomtitle=2mm,
    %     left=4mm
    % ]
    % \small
    % \texttt{You are a highly intelligent data extraction and reasoning agent. Your task is to analyze a conversation between a user and a tool/code-using AI model and convert it into a structured JSON format. You must adhere to the following principles:}
    % \begin{enumerate}
    %     \item \textbf{Semantic Action Inference:} \texttt{For each code block or bounding box, your primary goal is to infer its {intent}, not just its literal function. Crucially, this field must describe the question the model is trying to answer with the code/box, not the answer or observation it finds.}
    %     \item \textbf{Strict JSON Adherence:} \texttt{Your final output must be a single, valid JSON object... Do not include any explanations...}
    % \end{enumerate}
    % \end{tcolorbox}
    
    % \vspace{2mm} % Add some vertical space between the boxes

    % User Prompt Box
    \begin{tcolorbox}[
        colback=green!5,
        colframe=green!60!black,
        width=\textwidth,
        arc=2mm,
        auto outer arc,
        title=\textbf{System Prompt},
        before upper={\small},  % <--- 将字体命令作为选项
        fonttitle=\bfseries,
        boxsep=1mm,
        toptitle=1mm,
        bottomtitle=1mm,
        left=4mm
    ]
    \scriptsize
    \texttt{You are a helpful assistant. 
    Your goal is to solve the problem in the provided image(s) based on the user's instruction. Proceed step by step, optionally using the zoom-in tool one or more times to examine key areas closely. Selected regions will be cropped and processed externally, then re-encoded with your query to extract critical details.}

    \textbf{Tools}
    
    If needed, use the zoom-in tool one or more times to examine specific areas in detail.

    \textbf{Tool Format}
    
    Structure
    
    \begin{verbatim}
    {
        "region": [
            {
                "index": int, # Target image index to zoom in (0-based)
                "bbox_2d": list, # Format: [x1, y1, x2, y2], where (x1, y1) 
                is top-left corner and (x2, y2) is bottom-right corner
            },
            ... # Additional regions (optional)
        ],
        "query": str # Description of what to look for in the selected regions
    }

        \end{verbatim}

    \textbf{Parameters:}
    \begin{itemize}
        \item \texttt{region: List of dictionaries, each containing:}
        \begin{itemize}
            \item \texttt{index: Integer, specifying which image to zoom in}
            \item \texttt{bbox\_2d: List of $4$ integers [x1, y1, x2, y2] defining the region}
        \end{itemize}
        \item \texttt{query: String describing the search target}
    \end{itemize}
    
    \textbf{Constraints:}
    \begin{itemize}
        \item \texttt{At least one region must be specified}
        \item \texttt{All coordinates must be within image boundaries}
        \item \texttt{x1 < x2 and y1 < y2 must be satisfied}
    \end{itemize}

    \textbf{Example:}
    \begin{verbatim}
<tool> {"region": [{"index": 0, "bbox_2d": [100, 200, 300, 400]}], "query": "Look for the red button"} </tool>
    \end{verbatim}
    \end{tcolorbox}

    \caption{
        System prompt used in \textsc{ViLaVT} }
    \label{fig:sys_prompt}
\end{figure*}

\begin{figure*}[t!]
    \centering

    % User Prompt Box
    \begin{tcolorbox}[
        colback=green!5,
        colframe=green!60!black,
        width=\textwidth,
        arc=2mm,
        auto outer arc,
        title=\textbf{User Prompt},
        before upper={\small},  % <--- 将字体命令作为选项
        fonttitle=\bfseries,
        boxsep=1mm,
        toptitle=1mm,
        bottomtitle=1mm,
        left=4mm
    ]
    \scriptsize
    
The index of the provided image is 1\\
......\\
The index of the provided image is \textcolor{red}{\{max\_index\} }\\
\\
These are \textcolor{red}{\{{n\_frames}\} } images with indexed from 0 to \textcolor{red}{ \{max\_index\} }. \\
All images have size: width \textcolor{red}{\{width\}}, height \textcolor{red}{\{height\}}.\\
\\
\textbf{Question:} \textcolor{red}{ \{question\} } \\
\\
\textbf{If you need to zoom in for more details or examine specific regions, make tool call following the format:}

\begin{verbatim}
<think> Your reasoning about where to look and why  </think>
<tool> \{\{"region": [\{\{"index": int, "bbox\_2d": [x1, y1, x2, y2]\}\}, ...], "query": str\}\} </tool>
\end{verbatim}

\textbf{If you have enough information to answer the original question:}

\begin{verbatim}
<think> Your reasoning here.  </think>
<answer> Your final answer here. </answer>
\end{verbatim}

\begin{itemize}
    \item \texttt{Note that x1, y1, x2, y2 are the coordinates of the bounding box in the specfied image by the index.}
    \item \texttt{You must strictly follow the above output format.}
    \item \texttt{In $\langle$answer$\rangle$, provide \textbf{only} the final answer in the simplest required form:}
    \begin{itemize}
        \item \texttt{For multiple-choice questions: output only the letter (e.g., $A$, $B$, $C$).}
        \item \texttt{For yes/no questions: output only $Yes$ or $No$.}
        \item \texttt{For numerical answers: output only the number (e.g., $42$, $3.14$).}
        \item \texttt{Do not include explanations, units, punctuation, or extra words.}
    \end{itemize}
\end{itemize}
    \end{tcolorbox}

    \caption{
        User prompt used in \textsc{ViLaVT} }
    \label{fig:user_prompt}
\end{figure*}